\documentclass[twocolumn,10pt]{asme2ej}

\usepackage{graphicx} 
\usepackage{picinpar}
\usepackage{amsmath}
\usepackage{stfloats}
\usepackage{url}
\usepackage{flushend}
\usepackage{colortbl}
\usepackage{xcolor,soul,framed} 
\colorlet{shadecolor}{yellow}
\usepackage{multirow}
\usepackage{pifont}
\usepackage{color,soul}
\usepackage{alltt}
\usepackage[hidelinks]{hyperref}
\usepackage{enumerate}
\usepackage{siunitx}
\usepackage{breakurl}
\usepackage{epstopdf}
\usepackage{pbox}
\usepackage{amssymb}
\usepackage{comment}
\newtheorem{theorem}{Theorem}

\usepackage{subfigure}
\usepackage{algorithm}
\usepackage{algpseudocode}

\usepackage{autobreak}
\usepackage[percent]{overpic} 
\usepackage{rotating}
\graphicspath{{../figures/}}
\DeclareGraphicsExtensions{.eps,.eps_tex,.pdf,.jpeg,.png}
\definecolor{limegreen}{rgb}{0.2, 0.8, 0.2}
\definecolor{forestgreen}{rgb}{0.13, 0.55, 0.13}
\definecolor{greenhtml}{rgb}{0.0, 0.5, 0.0}
\definecolor{skyblue}{rgb}{0.53, 0.81, 0.92}
\definecolor{lightgray}{rgb}{0.83, 0.83, 0.83}
\definecolor{gray}{rgb}{0.75, 0.75, 0.75}
\definecolor{darkgray}{rgb}{0.66, 0.66, 0.66}
\usepackage{mathtools}
\colorlet{shadecolor}{yellow!255}
\renewenvironment{nomenclature}[1][1cm]{%
    \newcommand\entry[2]{\item[##1]##2\par}
    \section*{NOMENCLATURE}
    \list{}{\leftmargin #1}%
  }%
  {\endlist\par\addvspace{12pt}}

\title{Modeling, Vibration Control, and Trajectory Tracking of a Kinematically Constrained Planar Hybrid Cable-Driven Parallel Robot}

\author{Ronghuai Qi\thanks{Corresponding author.}\\
    \affiliation{
    Department of Mechanical and \\Mechatronics Engineering,\\
    University of Waterloo,\\
    Waterloo, ON N2L 3G1, Canada\\
    e-mail: ronghuai.qi@uwaterloo.ca
    }	
}

\author{Amir Khajepour\\
    \affiliation{
    Department of Mechanical and \\Mechatronics Engineering,\\
    University of Waterloo,\\
    Waterloo, ON N2L 3G1, Canada\\
    e-mail: a.khajepour@uwaterloo.ca
    }
}

\author{William W. Melek\\
    \affiliation{
    Department of Mechanical and \\Mechatronics Engineering,\\
    University of Waterloo,\\
    Waterloo, ON N2L 3G1, Canada\\
    e-mail: william.melek@uwaterloo.ca
    }
}

\begin{document}

\maketitle

\begin{abstract}
{\it This paper presents a kinematically constrained planar hybrid cable-driven parallel robot (HCDPR) for warehousing applications as well as other potential applications such as rehabilitation. The proposed HCDPR can harness the strengths and benefits of serial and cable-driven parallel robots. Based on this robotic platform, the goal in this paper is to develop an integrated control system to reduce vibrations and improve the trajectory accuracy and performance of the HCDPR, including deriving kinematic and dynamic equations, proposing solutions for redundancy resolution and optimization of stiffness, and developing two motion and vibration control strategies (controllers I and II). Finally, different case studies are conducted to evaluate the control performance, and the results show that the controller II can achieve the goal better.}
\end{abstract}


\begin{nomenclature}
\entry{${m_j}$}{Mass of the $j$th $(j=1,2)$ link of the robot arm.}
\entry{${m_m}$}{Mass of the mobile platform.}
\entry{${I_m}$}{Moment of inertia of the mobile platform.}
\entry{${I_j}$}{Moment of inertia of the $j$th $(j=1,2)$ link of the robot arm.}
\entry{${l_j}$}{Length of the $j$th $(j=1,2)$ link of the robot arm.}
\entry{${l_{cj}}$}{Length between the joint $j$ $(j=1,2)$ and the center of mass of link $j$ of the robot arm.}
\entry{${{\bf{p}}_e}$}{Vector of positions and orientation of the end-effector.} 
\entry{${\bf{q}}$}{Vector of generalized coordinates.} 
\entry{${\bf{\dot q}}$}{First order time-derivative of ${\bf{q}}$.}
\entry{${\bf{\ddot q}}$}{Second order time-derivative of ${\bf{q}}$.}
\entry{${{\bf{p}}_m}$}{Vector of positions and orientation of the mobile platform.} \entry{${\bf{\tau }}$}{Vector of generalized forces.} 
\entry{${\bf{T}}$}{Vector of cable tensions.} 
\entry{${{\bf{F}}_e}$}{External forces applied to the end-effector or the mobile platform.}
\entry{${{\bf{M}}_e}$}{External moment applied to the end-effector or the mobile platform.}
\entry{${g}$}{Gravitational acceleration.}
\entry{${{\bf{p}}_m}$}{The position and orientation of the mobile platform.}
\entry{${{\bf{A}}}$}{Structure matrix.}
\entry{${{\bf{L}}_i}$}{The position vector from the $i$th cable anchor point on the robot static frame to the $i$th cable anchor point on the mobile platform.}
\entry{${{\bf{\hat L}}_i}$}{Unit position vector of the $i$th cable.}
\entry{${\bf{L}}$}{Cable length vector.}
\entry{${\bf{L_0}}$}{Vector of unstretched cable lengths.}
\entry{${{\bf{T}}_i}$}{Vector of the $i$th cable tension.}
\entry{${\bf{K}}$}{Stiffness matrix.}
\entry{${{\bf{K}}_T}$}{Stiffness matrix as a product of the cable tensions.}
\entry{${{\bf{K}}_k}$}{Stiffness matrix as a product of the cable stiffness.}
\entry{${{\bf{K}}_c}$}{Cable stiffness matrix.}
\entry{$E_i$}{Elastic modulus of the $i$th cable.}
\entry{$A_{ci}$}{Cross section of the $i$th cable.}
\entry{${{\bf{J}}_e}$}{Jacobian matrix.}
\end{nomenclature}

\section{Introduction}
{S}{erial} manipulators are one of the most common types of industrial robots, which consist of a base, a series of links connected by actuator-driven joints, and an end-effector. Usually, they have 6 DOFs and offer high positioning accuracy. They are commonly used in industrial applications; however, they have some key limitations, such as high motion inertia and limited workspace envelope \cite{Wei2015}. {For example, the KUKA KR 60 HA is a 6-DOF serial robotic manipulator with a high payload ($60\;\rm{kg}$) carrying capacity and repeatability of $\pm 0.05\;{\rm{mm}}$, but the maximum reach is $2033\;{\rm{mm}}$ \cite{A.Dudarev2016}}. {CDPRs} are another important type of industrial robots. Their configurations usually bear resemblance to parallel manipulators (e.g., Stewart platform \cite{D.Stewart1965}). The NIST RoboCrane \cite{Albus1989,Albus1992} is a typical CDPR with 6 DOFs, which is designed by utilizing the idea of the Stewart platform, and its unique feature is its use of six cables instead of linear actuators. For these robots, rigid links are replaced with cables. This reduces the robot weight since cables are almost massless. It also eliminates the use of revolute joints. These features allow the mobile platform to reach high motion accelerations in large workspaces. For instance, some existing CDPRs were designed and analyzed in~\cite{Hiller2005,S.H.Yeo2013,Khajepour2015,Mendez2014}. However, they are not without some drawbacks, such as their low accuracy, high vibrations, etc., all of which limit their applications \cite{Oh2005}. To overcome the aforementioned shortages of serial and cable-driven parallel robots as well as aggregate their advantages, one approach is to combine these two types of robots to create a hybrid cable-driven parallel robot (HCDPR).

Some research and applications have been developed as follows: Albus~\cite{J.S.Albus1989} developed a cable-driven manipulator, where a robot arm was fixed upside down to the bottom of the RoboCrane robot{'}s platform~\cite{Albus1989,Albus1992} for lifting a load. Cable-driven camera systems are another type of cable-driven robots (cameras are affixed to the CDPRs) that can be used for different applications such as overhead filming~\cite{T.L.VINCENT2008}. {Gouttefarde}~\cite{Gouttefarde2016} developed a CDPR (called CoGiRo CSPR) with the onboard Yaskawa-Motoman SIA20 robot arm for contactless and interacting applications (e.g., spray painting and metal cutting), but this project still has the main problems of low stiffness of the CDPR which will result in vibrations.

However, the literature shows that existing research and applications prefer to affix a robot arm upside down to the bottom of a CDPR\textrm{'}s platform \cite{T.Arai1999,H.Osumi2000,M.Bamdad2015,M.Gouttefarde2017,J.S.Albus1989,J.S.Albus2003} or mainly control the cable robot while treating the serial robot as a manipulation tool or an end-effector rather than a whole system \cite{J.S.Albus1989,J.S.Albus2003}. When a serial robot is mounted on a mobile platform, the two constitute a new coupled system. Only controlling the mobile platform (i.e., treating the serial robot as a manipulation tool) or the serial robot may not guarantee the position accuracy of the end-effector. For applications that use such a system, the main goal is to control the end-effector of the serial robot (e.g., its trajectories and vibrations) in order to effectively accomplish tasks such as pick-and-place. Another major challenge in the utilization of these systems is maintaining the appropriate cable tensions and stiffness for the robot. {\color{black}The stiffness of CDPRs is an important issue, because driven cables are flexible, which reduce the robotic overall stiffness of the robots (compared to rigid cables) and produce undesired vibrations. When a CDPR moves, driven cables should maintain enough tensions to reduce vibrations, i.e., keep enough stiffness for the robot \cite{Behzadipour2006}. Regarding stiffness problem, some research has been developed: Behzadipour and Khajepour \cite{Behzadipour2006} have proposed an equivalent four-spring model to express the stiffness matrices of a CDPR. They also used a simulation example to verify this model. Azadi et al.~\cite{Azadi2009} introduced variable stiffness elements using antagonistic forces. Gosselin \cite{Gosselin1990} analyzed the stiffness mapping for parallel manipulators by considering the internal forces; conversely, Griffis and Duffy \cite{Griffis1993} modeled the global stiffness of a class of simple compliant couplings without considering the internal forces.} {While for a HCDPR, the moving robot arm also generates reaction forces acting on the mobile platform, resulting in mobile platform vibrations. Hence, it is challenging to achieve the goal of minimizing the vibrations and increasing the position accuracy of the end-effector simultaneously.} To the best of our knowledge, limited studies address the modeling and control problems of flexible HCDPRs, especially, when the redundancy and stiffness optimization problems are introduced, the control of trajectories and vibrations becomes more challenging. Although the research in \cite{Gouttefarde2016} showed a CDPR carrying a robot arm for painting large surfaces, vibrations were obvious and large based on their demonstration.

To solve the aforementioned problems in HCDPRs, the goal of this paper is to develop an integrated control system for the HCDPR to reduce vibrations and improve accuracy and performance. To achieve this goal, the following tasks are pursued: 1) derive analytical kinematic and dynamic equations for the HCDPR; 2) propose solutions for redundancy resolution and optimization of stiffness; 3) develop motion and vibration control methods for the HCDPR; and 4) conduct simulations to validate the effectiveness of the control methods proposed in Step 3). Additionally, the main contributions are as follows:
\begin{enumerate}
    \item A kinematically constrained planar HCDPR is proposed to harness the strengths and benefits of serial and cable-driven parallel robots. Detailed kinematic and dynamic equations are derived for this robot. An equivalent dynamic modeling (EDM) method is proposed to derive the dynamic equations of the HCDPR. This method has some advantages, e.g., by providing an effective solution for different configurations of HCDPRs. 
    \item Based on the configuration and models of the HCDPR, the redundancy resolution and stiffness maximization algorithms are proposed.
    \item Control strategies are designed for the HCDPR system to reduce vibrations and trajectory tracking errors. Compared to the existing study in~\cite{R.Qi2019j3,R.Qi2019j4}, this paper emphasizes the reaction performance between the mobile platform and the robot arm as well as trajectory tracking of the end-effector.
\end{enumerate}

Furthermore, the e-commerce explosion in recent years~\cite{Hamedthesis2018} stimulates the growth of automated warehousing solutions. By 2024, the market of global automated material handling equipment is predicted to no less than US\$ 50.0 Billion with a CAGR of $8\%$~\cite{MarketEngine2018}. These increase of automated warehousing applications offers a unique opportunity for the development of cable-driven robots. This paper provides a valid solution for these robots.

The rest of the paper is organized as follows: system modeling for the HCDPR is proposed in~\autoref{sec:J5_SystemModeling}. In~\autoref{sec:J5_ControlDesign}, the methods for redundancy resolution, optimization of stiffness, and controller design are proposed. Simulation results are evaluated in~\autoref{sec:J5_NumericalResults}. Finally, in~\autoref{sec:J5_Conclusions}, conclusions are summarized.

\section{Modeling of the Kinematically Constrained Planar HCDPR}\label{sec:J5_SystemModeling}
\subsection{HCDPR Configuration}\label{subsec:J5_HCDPRConfig}
\begin{figure}[!t]\centering
	\includegraphics[width=85mm]{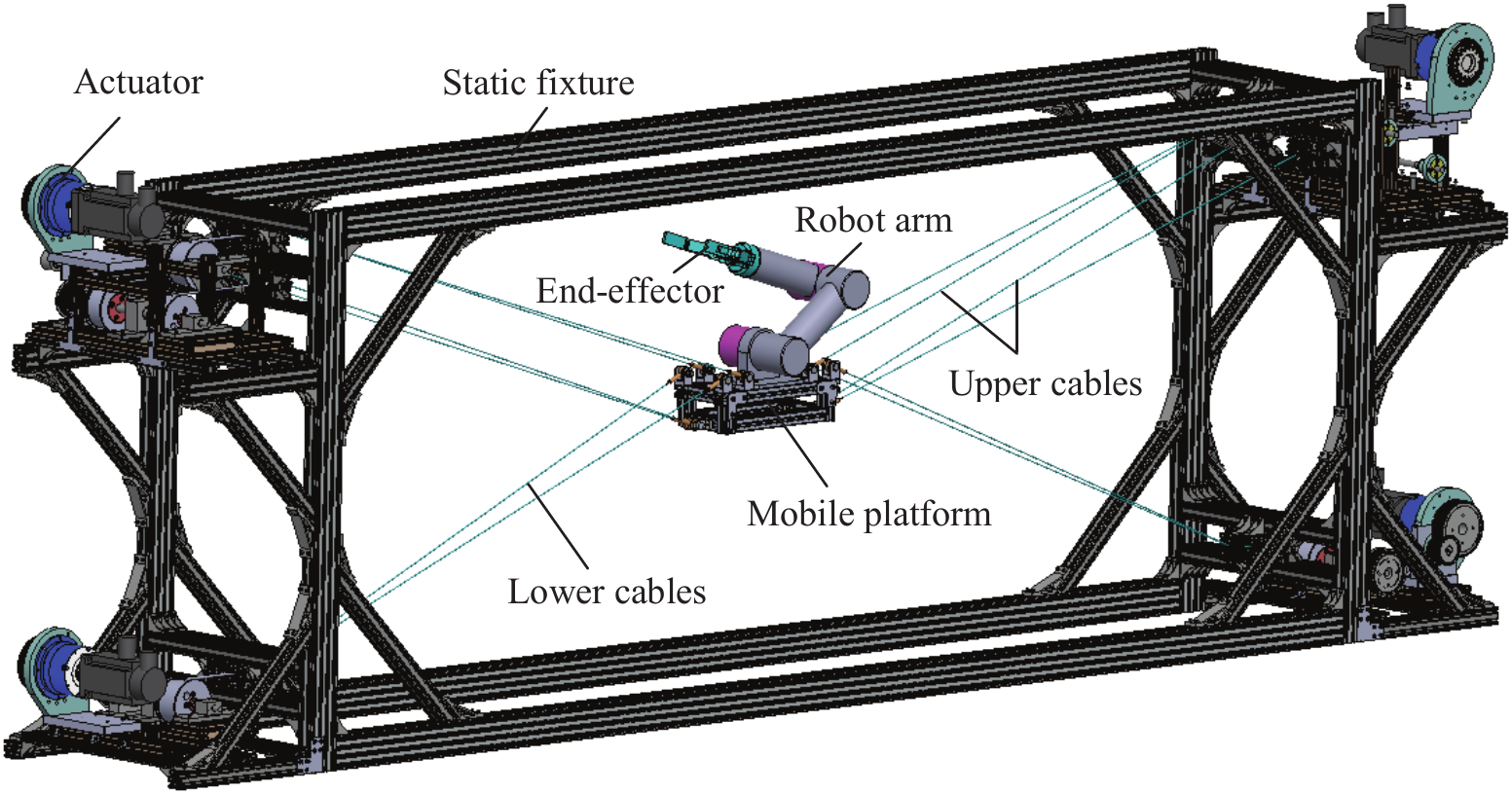}
	\caption{Mechanical model of the kinematically constrained planar HCDPR.}\label{fig:J5_MechModel}
\end{figure}

CDPRs are very useful in industries (e.g., warehousing), but they have some limitations (e.g., accuracy and vibrations). Serial robots have higher accuracy but are limited in terms of their workspace envelope. In this paper, the CDPR designed and studied in~\cite{Khajepour2015,Mendez2014,Rushton2016,R.Qi2018j2} is used for the development of the HCDPR, its integrated controller and evaluation and validation of the results. The proposed planar HCDPR consists of a 2-DOF robot arm (in the mechanical model shown in~\autoref{fig:J5_MechModel}), a 6-DOF rigid mobile platform, twelve cables, and four servo motors. The actuators are used to drive the cables to move the mobile platform. The robot arm is fixed on the mobile platform and moves with it. The twelve cables include four sets of cables: two sets of four-cable arrangement on the top and two sets of two-cable arrangement on the bottom. Each set of cables is controlled by one motor. In addition, the top actuators and bottom actuators control the upper cable lengths and lower cable tensions, respectively. The upper cables also restrict the orientation of the mobile platform, i.e., the kinematic constraints. The HCDPR parameters are shown in~\autoref{fig:J5_PlanarModel} and~\autoref{table:J5_HCDPRParameters}, respectively. The eight top cables and four bottom cables are simplified into four cables and two cables, respectively. The inertial coordinate frame $O\left\{ {{x_0},{y_0},{z_0}} \right\}$ is located at the center of the static fixture.

\subsection{HCDPR Kinematics}\label{subsec:J5_HCDPRKinematics}
\begin{figure}[!t]\centering
	\includegraphics[width=86mm]{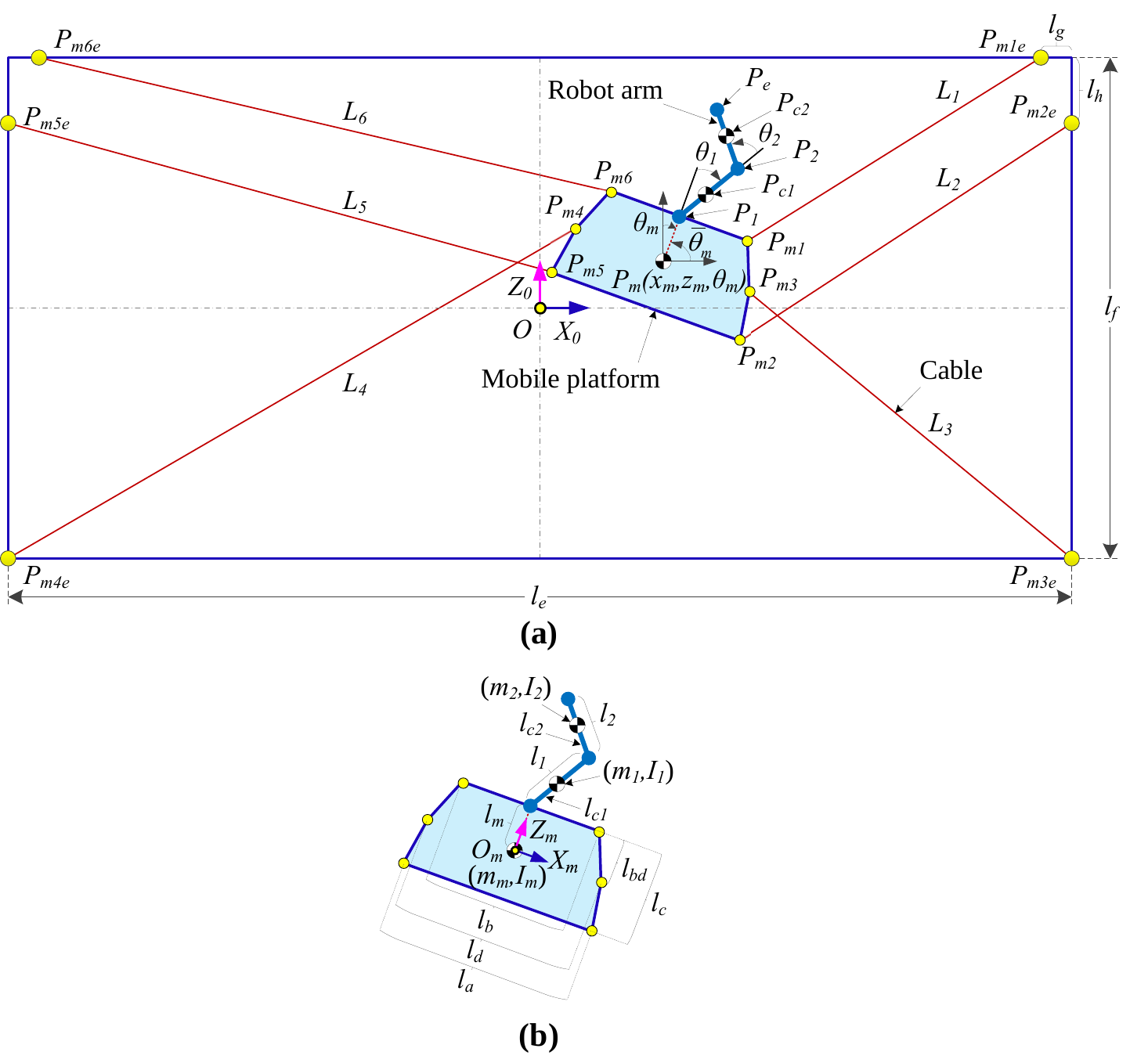}
	\caption{Configuration of the HCDPR. (a) planar HCDPR; (b) enlarged figure of the mobile platform with the robot arm.}\label{fig:J5_PlanarModel}
\end{figure}

The kinematics of the HCDPR includes forward kinematics and inverse kinematics. In this paper, analytical solutions of the kinematics will be derived for the planar HCDPR shown in~\autoref{fig:J5_PlanarModel}.

In \autoref{fig:J5_PlanarModel}, the planar HCDPR has 5 DOFs (the mobile platform has 3 DOFs and the robot arm has 2 DOFs), where point $P_m(x_m,z_m,\theta_m)$ is located at the center of mass of the mobile platform (indicating the degrees of freedom of the mobile platform); point $P_1(x_1,z_1)$ is located at the first joint of the attached robot arm; point $P_{c1}(x_{c1},z_{c1})$ represents the center of mass of robot link 1; point $P_2(x_2,z_2)$ is located at the second joint of the robot arm; point $P_{c2}(x_{c2},z_{c2})$ denotes the center of mass of robot link 2; and point $P_e(x_e,z_e,q_e)$ represents the positions and orientation of the robot end-effector. In addition, ${x_m}$ and ${z_m}$ denote the positions of the mobile platform (the center of mass) in the X-direction and Z-direction, respectively, with respect to the inertial coordinate frame $\{O\}$. {\color{black}${\theta_m}$ indicates the orientation of the mobile platform with respect to the $OY_0$-axis}, ${\color{black}\theta_{1}}$ and ${\color{black}\theta_{2}}$ are the relative angles of the robot arms as shown in \autoref{fig:J5_PlanarModel}. Other parameters of the HCDPR are also shown in~\autoref{fig:J5_PlanarModel} and~\autoref{table:J5_HCDPRParameters}.

The forward kinematics is derived by given the vector of generalized coordinates (joint variables) ${\bf{q}} = {[{x_m},{z_m},{\theta _m},{\color{black}\theta_{1}},}{{\color{black}\theta_{2}}]^T} \in {\mathbb{R}^5}$ to find the vector of positions and orientation ${{\bf{p}}_e} = {[{x_e},{z_e},{q_e}]^T} \in {\mathbb{R}^3}$ and the cable length vector ${\bf{L}} = {[{L_1},{L_2}, \cdots ,{L_6}]^T} \in {\mathbb{R}^6}$ as follows:
\begin{align}
{P_1}:&\left\{ \begin{array}{l}
{x_1} = {x_m} + {l_m}\cos \left( {{\bar \theta_m} } \right)\\
{z_1} = {y_m} + {l_m}\sin \left( {{\bar \theta_m} } \right)
\end{array} \right.\label{eq:J5_1}
\end{align}
\begin{align}
{P_{c1}}:&\left\{ \begin{array}{l}
{x_{c1}} = {x_1} + {l_{c1}}\cos ({\bar \theta_m}  + {\theta_{1}})\\
{z_{c1}} = {z_1} + {l_{c1}}\sin ({\bar \theta_m}  + {\theta_{1}})
\end{array} \right.\label{eq:J5_2}
\end{align}
\begin{align}
{P_2}:&\left\{ \begin{array}{l}
{x_2} = {x_1} + {l_1}\cos ({\bar \theta_m}  + {\theta_{1}})\\
{z_2} = {z_1} + {l_1}\sin ({\bar \theta_m}  + {\theta_{1}})
\end{array} \right.\label{eq:J5_3}
\end{align}
\begin{align}
{P_{c2}}:&\left\{ \begin{array}{l}
{x_{c2}} = {x_2} + {l_{c2}}\cos ({\bar \theta_m}  + {\theta_{1}} + {\theta_{2}})\\
{z_{c2}} = {z_2} + {l_{c2}}\sin ({\bar \theta_m}  + {\theta_{1}} + {\theta_{2}})
\end{array} \right.\label{eq:J5_4}
\end{align}
where ${\bar \theta_m} = {\theta_m} + \pi /2$, $(x_1,z_1)$, $(x_{c1},z_{c1})$, $(x_2,z_2)$, and $(x_{c2},z_{c2})$ represent the positions of joint 4, the center of mass of link 1 of the robot arm, joint 5, and the center of mass of link 2 of the robot arm, respectively.

\begin{table}[!t]
\renewcommand{\arraystretch}{1.3}
\caption{Parameters of the HCDPR}
\centering
\label{table:J5_HCDPRParameters}
\centering
	\begin{tabular}{c c c c}
	\hline\hline \\[-3mm]
    Symbol & Values & Symbol & Values  \\[1.6ex] \hline
    $l_{a}$ & $0.440$ \si{\metre} & $l_{bd}$ & $0.055$ \si{\metre} \\
    $l_{b}$ & $0.268$ \si{\metre} & $l_{g}$ & $0.086$ \si{\metre} \\
    $l_{c}$ & $0.105$ \si{\metre} & $l_{h}$ & $0.105$ \si{\metre} \\
    $l_{d}$ & $0.412$ \si{\metre} & $l_{m}$ & $0.052$ \si{\metre} \\
    $l_{e}$ & $3.000$ \si{\metre} & $l_{1}$ & $0.305$ \si{\metre} \\
    $l_{f}$ & $1.000$ \si{\metre} & $l_{2}$ & $0.305$ \si{\metre} \\
    $l_{c1}$ & $0.1525$ \si{\metre} & $l_{c2}$ & $0.1525$ \si{\metre} \\
    $m_m$ & \SI{30}{\kilogram} & $I_m$ & $0.83$ \si{\kilogram\metre\squared}\\
    $m_1$ & \SI{10}{\kilogram} & $I_1$ & $0.18$ \si{\kilogram\metre\squared}\\
    $m_2$ & \SI{10}{\kilogram} & $I_2$ & $0.18$ \si{\kilogram\metre\squared}\\
    $T_{min}$ & \SI{40}{\newton} & $T_{max}$ & \SI{2000}{\newton}\\
    $K_{s}$ & \SI{1.1e4}{\newton} & $g$ & $9.810$ \si[per-mode=symbol]{\metre\per\second\squared}\\
	\hline\hline
	\end{tabular}
\end{table}

The end-effector positions and orientation $(x_e,z_e,q_e)$ are described as
\begin{align}
{P_e}:\left\{ \begin{array}{l}
{x_e} = {x_2} + {l_2}\cos ({\bar \theta_m}  + {\theta_{1}} + {\theta_{2}})\\
{z_e} = {x_2} + {l_2}\sin ({\bar \theta_m}  + {\theta_{1}} + {\theta_{2}})\\
{q_e} = {\bar \theta_m}  + {\theta_{1}} + {\theta_{2}}
\end{array} \right..
\label{eq:J5_5}
\end{align}

The velocities of the center of mass of link 1 and link 2 are calculated as
\begin{align}
&\left\{ \begin{array}{l}
{{\dot x}_{c1}} = {{\dot x}_m} - {l_{c1}}({{\dot \theta}_m} + {{\dot \theta}_{1}})\sin ({\bar \theta_m}  + {\theta_{1}})\\
{\qquad\;}- {l_m}{{\dot \theta}_m}\sin \left( {{\bar \theta_m} } \right)\\
{{\dot z}_{c1}} = {{\dot y}_m} + {l_{c1}}({{\dot \theta}_m} + {{\dot \theta}_{1}})\cos ({\bar \theta_m}  + {\theta_{1}})\\
{\qquad\;}+ {l_m}{{\dot \theta}_m}\cos \left( {{\bar \theta_m} } \right)\\
{v_{c1}} = {\left( {\dot x_{c1}^2 + \dot z_{c1}^2} \right)^{1/2}}
\end{array} \right. \label{eq:J5_6}\\
&\left\{ \begin{array}{l}
{{\dot x}_{c2}} = {{\dot x}_m} - {l_1}({{\dot \theta}_m} + {{\dot \theta}_{1}})\sin ({\bar \theta_m}  + {\theta_{1}})\\
{\qquad\;}- {l_m}{{\dot \theta}_m}\sin({\bar \theta_m} ) - {l_{c2}}({{\dot \theta}_m} + {{\dot \theta}_{1}} + {{\dot \theta}_{2}})\\
{\qquad\;}\sin ({\bar \theta_m}  + {\theta_{1}} + {\theta_{2}})\\
{{\dot z}_{c2}} = {{\dot y}_m} + {l_{c1}}({{\dot \theta}_m} + {{\dot \theta}_{1}})\cos ({\bar \theta_m}  + {\theta_{1}})\\
{\qquad\;}+ {l_m}{{\dot \theta}_m}\cos({\bar \theta_m} ) + {l_{c2}}({{\dot \theta}_m} + {{\dot \theta}_{1}} + {{\dot \theta}_{2}})\\
{\qquad\;}\cos ({\bar \theta_m}  + {\theta_{1}} + {\theta_{2}})\\
{v_{c2}} = {\left( {\dot x_{c2}^2 + \dot z_{c2}^2} \right)^{1/2}}
\end{array} \right. \label{eq:J5_7}
\end{align}
where $({\dot x_{c1}},{\dot z_{c1}})$ and $({\dot x_{c2}},{\dot z_{c2}})$ represent the velocities the center of mass of link 1 and link 2 in the X-direction and Z-direction, respectively. ${v_{c1}}$ and ${v_{c2}}$ denote the total velocities of $({\dot x_{c1}},{\dot z_{c1}})$ and $({\dot x_{c2}},{\dot z_{c2}})$, respectively.

The Jacobian matrix ${{\bf{J}}_e}$ is calculated as
\begin{align}
{{\bf{J}}_e} = \frac{{d{{\bf{P}}_e}}}{{d{\bf{q}}}} = \left[\begin{matrix}
1&0&\begin{array}{l}
 - {l_1}{\mkern 1mu} \sin \left( {{\bar \theta_m}  + {\theta_{1}}} \right) - {l_m}{\mkern 1mu} \sin \left( {{\bar \theta_m} } \right)\\
 - {l_2}{\mkern 1mu} \sin \left( {{\bar \theta_m}  + {\theta_{1}} + {\theta_{2}}} \right)
\end{array}\\
0&1&\begin{array}{l}
{l_1}{\mkern 1mu} \cos \left( {{\bar \theta_m}  + {\theta_{1}}} \right) + {l_m}{\mkern 1mu} \cos \left( {{\bar \theta_m} } \right)\\
 + {l_2}{\mkern 1mu} \cos \left( {{\bar \theta_m}  + {\theta_{1}} + {\theta_{2}}} \right)
\end{array}\\
0&0&1
\end{matrix}\right.\nonumber\\
\left.\begin{matrix}
\begin{array}{l}
 - {l_1}{\mkern 1mu} \sin \left( {{\bar \theta_m}  + {\theta_{1}}} \right)\\
 - {l_2}{\mkern 1mu} \sin \left( {{\bar \theta_m}  + {\theta_{1}} + {\theta_{2}}} \right)
\end{array}&{ - {l_2}{\mkern 1mu} \sin \left( {{\bar \theta_m}  + {\theta_{1}} + {\theta_{2}}} \right)}\\
\begin{array}{l}
{l_1}{\mkern 1mu} \cos \left( {{\bar \theta_m}  + {\theta_{1}}} \right)\\
 + {l_2}{\mkern 1mu} \cos \left( {{\bar \theta_m}  + {\theta_{1}} + {\theta_{2}}} \right)
\end{array}&{{l_2}{\mkern 1mu} \cos \left( {{\bar \theta_m}  + {\theta_{1}} + {\theta_{2}}} \right)}\\
1&1
\end{matrix}\right].
\label{eq:J5_8}
\end{align}

For the cable length ${\bf{L}}$, first, the corresponding vectors shown in~\autoref{fig:J5_PlanarModel} are computed as
\begin{align}
\left\{ \begin{array}{l}
{{\bf{p}}_{m1e}} = {\begin{bmatrix}
{{l_e}/2 - {l_g}}&0&{{l_f}/2}
\end{bmatrix}^T}\\
{{\bf{p}}_{m2e}} = {\begin{bmatrix}
{{l_e}/2}&0&{{l_f}/2 - {l_h}}
\end{bmatrix}^T}\\
{{\bf{p}}_{m3e}} = {\begin{bmatrix}
{{l_e}/2}&0&{ - {l_f}/2}
\end{bmatrix}^T}\\
{{\bf{p}}_{m4e}} = {\begin{bmatrix}
{ - {l_e}/2}&0&{ - {l_f}/2}
\end{bmatrix}^T}\\
{{\bf{p}}_{m5e}} = {\begin{bmatrix}
{ - {l_e}/2}&0&{{l_f}/2 - {l_h}}
\end{bmatrix}^T}\\
{{\bf{p}}_{m6e}} = {\begin{bmatrix}
{ - {l_e}/2 + {l_g}}&0&{{l_f}/2}
\end{bmatrix}^T}
\end{array} \right.
\label{eq:J5_9}
\end{align}
and
\begin{align}
\left\{ \begin{array}{l}
{{\bf{p}}_{m1}} = {\begin{bmatrix}
{{\color{black}x_m}}&0&{{\color{black}z_m}}
\end{bmatrix}^T} + {\bf{R}_y}({{\color{black}\theta_m}}){\begin{bmatrix}
{{l_b}/2}&0&{{l_m}}
\end{bmatrix}^T}\\
{{\bf{p}}_{m2}} = {\begin{bmatrix}
{{\color{black}x_m}}&0&{{\color{black}z_m}}
\end{bmatrix}^T} + {\bf{R}_y}({{\color{black}\theta_m}}){\begin{bmatrix}
{{l_a}/2}&0&{{l_m} - {l_c}}
\end{bmatrix}^T}\\
{{\bf{p}}_{m3}} = {\begin{bmatrix}
{{\color{black}x_m}}&0&{{\color{black}z_m}}
\end{bmatrix}^T} + {\bf{R}_y}({{\color{black}\theta_m}}){\begin{bmatrix}
{{l_d}/2}&0&{{l_m} - {l_{bd}}}
\end{bmatrix}^T}\\
{{\bf{p}}_{m4}} = {\begin{bmatrix}
{{\color{black}x_m}}&0&{{\color{black}z_m}}
\end{bmatrix}^T} + {\bf{R}_y}({{\color{black}\theta_m}}){\begin{bmatrix}
{ - {l_d}/2}&0&{{l_m} - {l_{bd}}}
\end{bmatrix}^T}\\
{{\bf{p}}_{m5}} = {\begin{bmatrix}
{{\color{black}x_m}}&0&{{\color{black}z_m}}
\end{bmatrix}^T} + {\bf{R}_y}({{\color{black}\theta_m}}){\begin{bmatrix}
{ - {l_a}/2}&0&{{l_m} - {l_c}}
\end{bmatrix}^T}\\
{{\bf{p}}_{m6}} = {\begin{bmatrix}
{{\color{black}x_m}}&0&{{\color{black}z_m}}
\end{bmatrix}^T} + {\bf{R}_y}({{\color{black}\theta_m}}){\begin{bmatrix}
{ - {l_b}/2}&0&{{l_m}}
\end{bmatrix}^T}
\end{array} \right.
\label{eq:J5_10}
\end{align}
where ${{\bf{p}}_{mie}}\;\left( {i = 1,2, \cdots ,6} \right)$ and ${{\bf{p}}_{mi}}\;\left( {i = 1,2, \cdots ,6} \right)$ represent the position vectors of the $i$th cable anchor point on the robot static frame and the $i$th cable anchor point on the mobile platform, respectively. ${\bf{R}_y}({{\color{black}\theta_m}})$ is the rotation matrix along the Y-axis (moving frame).

Then, the cable position vector is calculated as
\begin{align}
{{\bf{L}}_i} = {{\bf{p}}_{mie}} - {{\bf{p}}_{mi}}\quad \quad i = 1,2, \cdots ,6.
\label{eq:J5_11}
\end{align}

Finally, the cable length vector is computed as
\begin{align}
{\bf{L}} &= {\begin{bmatrix}
{{L_1}}&{{L_2}}&{{L_3}}&{{L_4}}&{{L_5}}&{{L_6}}
\end{bmatrix}^T}\nonumber\\
&= {\begin{bmatrix}
{\left\| {{{\bf{L}}_1}} \right\|}&{\left\| {{{\bf{L}}_2}} \right\|}&{\left\| {{{\bf{L}}_3}} \right\|}&{\left\| {{{\bf{L}}_4}} \right\|}&{\left\| {{{\bf{L}}_5}} \right\|}&{\left\| {{{\bf{L}}_6}} \right\|}
\end{bmatrix}^T}.
\label{eq:J5_12}
\end{align}

The inverse kinematics is calculated by given the vector of positions and orientation ${{\mathbf{p}}_e} = {\begin{bmatrix}
  {{x_e}}&{{z_e}}&{{q_e}} 
\end{bmatrix}^T} \in {\mathbb{R}^3}$ and the cable length vector ${\mathbf{L}} = {\begin{bmatrix}
  {{L_1}}&{{L_2}}& \cdots &{{L_6}} 
\end{bmatrix}^T}$ to find the vector of joint variables ${\mathbf{q}} = \begin{bmatrix}
  {{\color{black}x_m}}&{{\color{black}z_m}}& \cdots &{{\color{black}\theta_{2}}} \end{bmatrix}^T$ as follows.

Suppose the cable lengths ${L_1}$ and ${L_6}$ are given and the kinematic constraints are applied (i.e., ${L_1} = {L_2}$ and ${L_5} = {L_6}$, then ${{\color{black}\theta_m}} = 0$). Then, the solutions of ${\color{black}x_m}$ and ${\color{black}z_m}$ are computed as
\begin{align}
\left\{ \begin{array}{l}
{\color{black}x_m} = \frac{{L_1^2 - L_6^2}}{{2({l_b} - {l_e} + 2{l_g})}}\\
{\color{black}z_m} = \frac{{{l_f}}}{2} - {l_c} + {l_m} \pm \frac{1}{{2\left( {{l_b} - {l_e} + 2{l_g}} \right)}}\\
\left( {\left( {{L_1} - {L_6} + {l_b} - {l_e} + 2{l_g}} \right)\left( {{L_6} - {L_1} + {l_b} - {l_e} + 2{l_g}} \right)} \right.\\
{\left. {\left( {{L_1} + {L_6} + {l_b} - {l_e} + 2{l_g}} \right)\left( {{L_1} + {L_6} - {l_b} + {l_e} - 2{l_g}} \right)} \right)^{\frac{1}{2}}}.
\end{array} \right.
\label{eq:J5_13}
\end{align}

Substituting ${\color{black}x_m}$, ${\color{black}z_m}$, and ${\color{black}\theta_m}$ into~\eqref{eq:J5_1}, we can find $(x_1, z_1)$. Other terms are calculated as
\begin{align}
\left\{ \begin{array}{l}
{r_{1e}}: = {\left( {{{\left( {{x_e} - {x_1}} \right)}^2} + {{\left( {{z_e} - {z_1}} \right)}^2}} \right)^{1/2}}\\
\phi : = {\rm{atan2}}\left( {{z_e} - {z_1},{x_e} - {x_1}} \right)\\
\alpha : = {\cos ^{ - 1}}\left( {\frac{{r_{1e}^2 + l_1^2 - l_2^2}}{{2{r_{1e}}{l_1}}}} \right)\\
\beta : = {\cos ^{ - 1}}\left( {\frac{{l_1^2 + l_2^2 - r_{1e}^2}}{{2{l_1}{l_2}}}} \right).
\end{array} \right.
\label{eq:J5_15}
\end{align}

Finally, two solutions for ${\color{black}\theta_{1}}$ and ${\color{black}\theta_{2}}$ are computed (using \eqref{eq:J5_15}) as
\begin{align}
\left\{ \begin{array}{l}
{\color{black}\theta_{1}} = \phi  \mp \alpha  - {{\color{black}\theta_m}}\\
{\color{black}\theta_{2}} =  \pm \left( {\pi  - \beta } \right)
\end{array} \right.
\label{eq:J5_16}
\end{align}
where \eqref{eq:J5_15} and \eqref{eq:J5_16} are available whether the cable kinematic constraints are applied or not.

%
\subsection{HCDPR Dynamics}\label{subsec:J5_HCDPRDynamics}
In this paper, a method called equivalent dynamic modeling (EDM) is used to derive the dynamic equations of the HCDPR using the following steps: 
1)	mapping cable tensions ($n$ cables) to the equivalent joint forces/torques ($k$ DOFs) for the cable-driven robot;
2)	derive equivalent $m$-DOF robot dynamic equations (the attached robot arm has ($m-k$) DOFs) which include equivalent joint forces/torques ($k$ DOFs);
3)	express the corresponding terms of equivalent joint forces/torques (in the equivalent $m$-DOF robot dynamic equations obtained in Step 2) in terms of the n cable tensions. Then, the analytical dynamic model of the HCDPR will be introduced. The EDM method has some advantages of deriving dynamic equations for HCDPRs. For example, it provides an effective solution for different configurations of HCDPRs. In this paper, the EDM method will be applied to the planar HCDPR shown in \autoref{fig:J5_PlanarModel}. The equivalent dynamic modeling method applied to the planar HCDPR is conducted as follows:

1)	An equivalent three-spring driven model shown in Appendix~\ref{appendix:J5_EquivalentModel} is developed. A cable-tension transformation equation ${{\bf{\tau }}_m} =  - {\bf{AT}}$ is satisfied and proved, where ${{\mathbf{\tau }}_m}: = {\left[ {{\tau _x},{\tau _z},{\tau _\theta }} \right]^T} \in {\mathbb{R}^3}$, ${\mathbf{A}} \in {\mathbb{R}^{3 \times n}},$ and ${\mathbf{T}}: = {\left[ {{T_1},{T_2}, \cdots ,{T_n}} \right]^T} \in {\mathbb{R}^n}$ represent the equivalent joint forces/torques applied to the mobile platform, the structure matrix $\bf{A}$, and the cable tensions, respectively. 

2)	The kinetic and potential energy are calculated as
\begin{align}
{K_E} = &{}\frac{1}{2}{m_m}\dot {\color{black}x}_m^2 + \frac{1}{2}{m_m}\dot {\color{black}z}_m^2 + \frac{1}{2}{I_m}\dot {\color{black}\theta}_m^2 + \frac{1}{2}{m_1}v_{c1}^2 \nonumber\\
&{}+ \frac{1}{2}{I_1}{\left( {{{\dot \theta}_m} + {{\dot \theta}_{1}}} \right)^2} + \frac{1}{2}{m_2}v_{c2}^2 + \frac{1}{2}{I_2}{\left( {{{\dot \theta}_m} + {{\dot \theta}_{1}} + {{\dot \theta}_{2}}} \right)^2}
\label{eq:J5_17}
\end{align}
and 
\begin{align}
{P_E} = &{}{m_m}g{\color{black}z_m} + {m_1}g{z_{c1}} + {m_2}g{z_{c2}} + \frac{1}{2}{k_x}{\left( {{\color{black}x_m} - {\color{black}x_{m0}}} \right)^2} \nonumber\\
&{}+ \frac{1}{2}{k_z}{\left( {{\color{black}z_m} - {\color{black}z_{m0}}} \right)^2} + \frac{1}{2}{k_\theta }{\left( {{{\color{black}\theta_m}} - {\color{black}\theta_{m0}}} \right)^2}
\label{eq:J5_18}
\end{align}
where ${k_x}$, ${k_z}$, and ${k_\theta }$ come from the equivalent three-spring driven model shown in \ref{appendix:J5_EquivalentModel}, which represent the corresponding spring constants based on Hooke's law. Also, the expression $\frac{1}{2}{k_x}{\left( {{\color{black}x_m} - {\color{black}x_{m0}}} \right)^2} + \frac{1}{2}{k_z}{\left( {{\color{black}z_m} - {\color{black}z_{m0}}} \right)^2} + \frac{1}{2}{k_\theta }{\left( {{{\color{black}\theta_m}} - {\color{black}\theta_{m0}}} \right)^2}$ indicates the spring potential energy of the equivalent joints.

Then, the Lagrange's equation is described as
\begin{align}
\frac{d}{{dt}}\left( {\frac{{\partial ({K_E} - {P_E})}}{{\partial {{\dot q}_j}}}} \right) - \frac{{\partial ({K_E} - {P_E})}}{{\partial {q_j}}} = {\tau _j}\quad j = 1,2, \cdots ,5.
\label{eq:J5_19}
\end{align}

The dynamic equation is computed as
\begin{align}
{\mathbf{M}}\left( {\mathbf{q}} \right){\mathbf{\ddot q + C}}\left( {{\mathbf{q}},{\mathbf{\dot q}}} \right){\mathbf{\dot q + G}}\left( {\mathbf{q}} \right) + {{\mathbf{P}}_{vs}}\left( {\mathbf{q}} \right)={\mathbf{\tau}}
\label{eq:J5_20}
\end{align}
where $\mathbf{q} \in {\mathbb{R}^{5}}$, ${\mathbf{\dot q}} \in \mathbb{R}{^{5}}$, and ${\mathbf{\ddot q}} \in \mathbb{R}{^{5}}$, represent the vectors of generalized coordinates, velocities, and accelerations, respectively. ${\mathbf{M}}({\mathbf{q}}) \in \mathbb{R}{^{5 \times 5}}$, $\mathbf{C}({{{\mathbf{q}}},{\mathbf{\dot q}}}) \in \mathbb{R}{^{5 \times 5}}$, $\mathbf{G}(\mathbf{q}) \in {\mathbb{R}^{5}}$, and ${\mathbf{P}}_{vs}(\mathbf{q}) \in {\mathbb{R}^{5}}$ denote the inertia matrix, Coriolis and centripetal matrix, vector of gravitational force, and vector of elastic force, respectively. ${\mathbf{\tau}} \in {\mathbb{R}^{5}}$ represents the vector of generalized force. ${\mathbf{M}}({\mathbf{q}}), \mathbf{C}({{{\mathbf{q}}},{\mathbf{\dot q}}}),\mathbf{G}(\mathbf{q})$, and ${{\mathbf{P}}_{vs}}(\mathbf{q})$ are provided in Appendix~\ref{appendix:J5_HCDPR_Drivations}.

When an external force ${{\bf{F}}_e}$ and moment ${{\bf{M}}_e}$ are applied to the end-effector, the dynamic equation can be rewritten as
\begin{align}
{\bf{M}}\left( {\bf{q}} \right){\bf{\ddot q + C}}\left( {{\bf{q}},{\bf{\dot q}}} \right){\bf{\dot q + G}}\left( {\bf{q}} \right) + {{\bf{P}}_{vs}}\left( {\bf{q}} \right){\bf{ + J}}_e^T{\begin{bmatrix}
{{{\bf{F}}_e}}&{{{\bf{M}}_e}}
\end{bmatrix}^T}={\bf{\tau}}
\label{eq:J5_21}
\end{align}
where ${{\bf{J}}_e}$ is the Jacobian matrix. Here, \eqref{eq:J5_20} or \eqref{eq:J5_21} is the equivalent HCDPR dynamic equation.

3)	For the planar HCDPR, \eqref{eq:J5_21} and~\eqref{eq:J5_appendix_A_4} are rearranged as
\begin{align}
&{\bf{M}}\left( {\bf{q}} \right){\bf{\ddot q + C}}\left( {{\bf{q}},{\bf{\dot q}}} \right){\bf{\dot q + G}}\left( {\bf{q}} \right) + {\bf{J}}_e^T{\begin{bmatrix}
{{{\bf{F}}_e}}&{{{\bf{M}}_e}}
\end{bmatrix}^T} \nonumber\\
&={\bf{\tau }} - {{\bf{P}}_{vs}}\left( {\bf{q}} \right) = \begin{bmatrix}
{{\tau _1} - {P_{vs1}}}\\
{{\tau _2} - {P_{vs2}}}\\
{{\tau _3} - {P_{vs3}}}\\
{{\tau _4}}\\
{{\tau _5}}
\end{bmatrix}.
\label{eq:J5_22}
\end{align}

Using \eqref{eq:J5_10}, \eqref{eq:J5_11}, and \eqref{eq:J5_12}, and by supposing ${\hat L_k} = \frac{{{L_k}}}{{\left\| {{{\mathbf{L}}_k}} \right\|}} \in {\mathbb{R}^3}$ and ${r_k} = \left( {{p_{mi}} - {p_m}} \right)$, where $i = 1,2, \cdots ,6$. Then, a matrix $\bf{A}$ is defined as
\begin{align}
{\bf{A}} = \begin{bmatrix}
{{{\hat L}_{1x}}}& \cdots &{{{\hat L}_{6x}}}\\
{{{\hat L}_{1z}}}& \cdots &{{{\hat L}_{6z}}}\\
{{{\begin{bmatrix}
{{r_{1x}}}\\{{r_{1z}}}
\end{bmatrix}}} \times {{\begin{bmatrix}
{{{\hat L}_{1x}}}\\{{{\hat L}_{1z}}}
\end{bmatrix}}}}& \cdots &{{{\begin{bmatrix}
{{r_{6x}}}\\{{r_{6z}}}
\end{bmatrix}}} \times {{\begin{bmatrix}
{{{\hat L}_{6x}}}\\{{{\hat L}_{6z}}}
\end{bmatrix}}}}
\end{bmatrix}.
\label{eq:J5_23}
\end{align}

Combining~\eqref{eq:J5_appendix_A_8} and \eqref{eq:J5_23}, $\bf{AT}$ is the force and moment (applied at the center of mass of the mobile platform) coming from the flexible cables. Based on the Lagrange's equation \eqref{eq:J5_19}, ${\tau _j}\;(j = 1,2, \ldots ,5)$ denotes the generalized force/torque applied to the dynamic system at joint j to drive link j, but in the specific HCDPR, the mobile platform is driven by six cables, i.e., there is no direct input (force/torque) applied to the center of mass of the mobile platform, so ${\tau _1} = {\tau _2} = {\tau _3} = 0$. From \eqref{eq:J5_22} and~\eqref{eq:J5_appendix_A_10}, we get
\begin{align}
&{\begin{bmatrix}
{{\tau _1} - {P_{vs1}}}&{{\tau _2} - {P_{vs2}}}&{{\tau _3} - {P_{vs3}}}
\end{bmatrix}^T} \nonumber\\
&= {\begin{bmatrix}
{ - {P_{vs1}}}&{ - {P_{vs2}}}&{ - {P_{vs3}}}
\end{bmatrix}^T} =  - {{\bf{\tau }}_m} = {\bf{AT}}.
\label{eq:J5_24}
\end{align}

Then, by combining \eqref{eq:J5_22} and \eqref{eq:J5_24}, the dynamic equation of HCDPR is described as
\begin{align}
{\bf{M}}\left( {\bf{q}} \right){\bf{\ddot q + C}}\left( {{\bf{q}},{\bf{\dot q}}} \right){\bf{\dot q + G}}\left( {\bf{q}} \right) + {\bf{J}}_e^T{\begin{bmatrix}
{{{\bf{F}}_e}}&{{{\bf{M}}_e}}
\end{bmatrix}^T}{\bf{ = }}\begin{bmatrix}
{{\bf{AT}}}\\
{{\tau _4}}\\
{{\tau _5}}
\end{bmatrix}.
\label{eq:J5_25}
\end{align}

In addition, consider the configuration and constraints of HCDPR shown in~\autoref{fig:J5_PlanarModel}. The upper cables and lower cables are based on position control and force control, respectively. Then, the cable tensions $\bf{T}$ shown in \eqref{eq:J5_25} are calculated as
\begin{align}
\left\{ \begin{array}{l}
{T_1} = \frac{{{K_s}}}{{{L_{01}}}}\left( {{L_1} - {L_{01}}} \right)\\
{T_2} = \frac{{{K_s}}}{{{L_{02}}}}\left( {{L_2} - {L_{02}}} \right)\\
{T_3} = {T_3}\\
{T_4} = {T_4}\\
{T_5} = \frac{{{K_s}}}{{{L_{05}}}}\left( {{L_5} - {L_{05}}} \right)\\
{T_6} = \frac{{{K_s}}}{{{L_{06}}}}\left( {{L_6} - {L_{06}}} \right)
\end{array} \right.
\label{eq:J5_26}
\end{align}
where the unstretched cable lengths are ${L_{01}} = {L_{02}}$ and ${L_{05}} = {L_{06}}$ (because of the kinematic constraints), ${L_i}$ $(i = 1,2, \cdots, 6)$ are the state of current cable lengths, and ${K_s}$ is the specific stiffness. Hence, suppose the inputs of the real HCDPR are ${\bf{u}} = {\begin{bmatrix}
{{L_{01}}}&{{T_3}}&{{T_4}}&{{L_{06}}}
\end{bmatrix}^T}$ and ${{\bf{\tau }}_{45}} = {\begin{bmatrix}
{{\tau _4}}&{{\tau _5}}
\end{bmatrix}^T}$ (torques applied to the revolute joint on the robot arm). Also, the outputs are assumed to be ${\bf{q}} = {\begin{bmatrix}
{{\color{black}x_m}}&{{\color{black}z_m}}&{{{\color{black}\theta_m}}}&{{\color{black}\theta_{1}}}&{{\color{black}\theta_{2}}}
\end{bmatrix}^T}$, where ${\color{black}x_m}$ and ${\color{black}z_m}$ are the equivalent prismatic joints on the mobile platform; ${\color{black}\theta_m}$ is the equivalent revolute joint on the mobile platform; and ${\color{black}\theta_{1}}$ and ${\color{black}\theta_{2}}$ are the revolute joints on the robot arm.

\section{Control Design}
\label{sec:J5_ControlDesign}
Based on the system modeling in~\autoref{sec:J5_SystemModeling}, the redundancy resolution, stiffness optimization problem, and controller design will be proposed to address vibration control and trajectory tracking issues.

%
\subsection{Redundancy Resolution}\label{subsec:J5_RedundancyResolution}
For the HCDPR shown in \autoref{fig:J5_PlanarModel}, the robot is redundant in terms of the number of degrees of freedom (i.e., six cables drive the 3-DOF mobile platform).  Suppose an external force and moment ${\left[ {{{\bf{F}}_e},{{\bf{M}}_e}} \right]^T}$ are applied to the center of mass of the mobile platform, we get
\begin{align}
{\mathbf{T}} = {{\mathbf{A}}^ + }\left( {{m_m}{{\begin{bmatrix}
  0&0&g 
\end{bmatrix}}^T} + {{\left[ {{{\mathbf{F}}_e},{{\mathbf{M}}_e}} \right]}^T}} \right) \in {\mathbb{R}^6}.
\label{eq:J5_3-1}
\end{align}

By supposing the wrench vector ${{\mathbf{W}}_m}: = {m_m}{\begin{bmatrix}
  0&0&g 
\end{bmatrix}^T} + {\left[ {{{\mathbf{F}}_e},{{\mathbf{M}}_e}} \right]^T} \in {\mathbb{R}^3}$, then
\begin{align}
{\bf{T}} = {{\bf{A}}^T}{\left( {{\bf{A}}{{\bf{A}}^T}} \right)^{ - 1}}{{\bf{W}}_m}.
\label{eq:J5_3-2}
\end{align}

In \eqref{eq:J5_3-2}, the elements of $\bf{T}$ (i.e., cable tensions) might be negative. However, in the real system, they cannot drive the mobile platform if they were negative. The redundancy resolution of the cable tensions T can be formulated as
\begin{align}
{\bf{T}} = {{\bf{A}}^T}{\left( {{\bf{A}}{{\bf{A}}^T}} \right)^{ - 1}}{{\bf{W}}_m} + {\rm{Null}}\left( {\bf{A}} \right){\begin{bmatrix}
{{\lambda _1}}&{{\lambda _2}}&{{\lambda _3}}
\end{bmatrix}^T}
\label{eq:J5_3-3}
\end{align}
where ${\mathbf{T}} = {\begin{bmatrix}
  {{T_1}}&{{T_2}}& \cdots &{{T_6}} 
\end{bmatrix}^T} \in {\mathbb{R}^6}$, ${\rm{Null}}\left( {\mathbf{A}} \right)$ represents the null space of structure matrix A (A is calculated using \eqref{eq:J5_23}, and ${\lambda _1},{\lambda _2},{\lambda _3} \in \mathbb{R}$ are three arbitrary values. In \eqref{eq:J5_3-3}, ${\rm{Null}}\left( {\bf{A}} \right){\begin{bmatrix}
{{\lambda _1}}&{{\lambda _2}}&{{\lambda _3}}
\end{bmatrix}^T}$ belongs to the null space of $\bf{A}$, since it can be described as ${\bf{A}}\left( {{\rm{Null}}\left( {\bf{A}} \right){{\begin{bmatrix}
{{\lambda _1}}&{{\lambda _2}}&{{\lambda _3}}
\end{bmatrix}}^T}} \right) = \left( {{\bf{A}}{\rm{Null}}\left( {\bf{A}} \right)} \right){\begin{bmatrix}
{{\lambda _1}}&{{\lambda _2}}&{{\lambda _3}}
\end{bmatrix}^T} = {\bf{0}}{\begin{bmatrix}
{{\lambda _1}}&{{\lambda _2}}&{{\lambda _3}}
\end{bmatrix}^T} = 0$. The expression ${\rm{Null}}\left( {\bf{A}} \right){\begin{bmatrix}
{{\lambda _1}}&{{\lambda _2}}&{{\lambda _3}}
\end{bmatrix}^T}$ denotes antagonistic cable tensions. The cable tensions T increase if all the antagonistic cable tensions are positive. Hence, the values of ${\lambda _1},{\lambda _2},{\lambda _3}$ can be selected to maintain that all the cable tensions are positive.

By supposing ${{\bf{T}}_A}: = {{\bf{A}}^T}{\left( {{\bf{A}}{{\bf{A}}^T}} \right)^{ - 1}}{{\bf{W}}_m}$ and ${{\bf{N}}_A}: = {\rm{Null}}\left( {\bf{A}} \right)$, then \eqref{eq:J5_3-3} is rearranged as
\begin{align}
{\bf{T}} = {{\bf{T}}_A} + {{\bf{N}}_A}{\begin{bmatrix}
{{\lambda _1}}&{{\lambda _2}}&{{\lambda _3}}
\end{bmatrix}^T}
\label{eq:J5_3-4}
\end{align}
where ${{\mathbf{T}}_A} = {\begin{bmatrix}
  {{T_{A1}}}&{{T_{A2}}}& \cdots &{{T_{A6}}} 
\end{bmatrix}^T} \in {\mathbb{R}^6}$ and ${{\mathbf{N}}_A} = \begin{bmatrix}
  {{N_{A11}}}&{{N_{A12}}}&{{N_{A13}}} \\ 
  {{N_{A21}}}&{{N_{A22}}}&{{N_{A23}}} \\ 
  {{N_{A31}}}&{{N_{A33}}}&{{N_{A33}}} \\ 
  {{N_{A41}}}&{{N_{A42}}}&{{N_{A43}}} \\ 
  {{N_{A51}}}&{{N_{A52}}}&{{N_{A53}}} \\ 
  {{N_{A61}}}&{{N_{A62}}}&{{N_{A63}}} 
\end{bmatrix} \in {\mathbb{R}^{6 \times 3}}$. Eq. \eqref{eq:J5_3-4} is introduced to combine the stiffness maximization method and constraints in order to optimize cable tensions.

%
\subsection{Maximizing Stiffness of the HCDPR}\label{subsec:J5_MaxStiffness}
To calculate the stiffness matrix ${\bf{K}}$ for a static cable-driven robot, first, suppose an external force and moment ${\left[ {{{\bf{F}}_e},{{\bf{M}}_e}} \right]^T}$ are applied to the center of mass of the mobile platform. The stiffness matrix ${\bf{K}}$ is computed as
\begin{align}
{\bf{K}}:=&{} \frac{{d\left( {{m_m}{{\begin{bmatrix}
0&0&g
\end{bmatrix}}^T} + {{\left[ {{{\bf{F}}_e},{{\bf{M}}_e}} \right]}^T}} \right)}}{{d{{\bf{p}}_m}}} = \frac{{d({\bf{AT}})}}{{d{{\bf{p}}_m}}}\nonumber\\
=&{} \frac{{d{\bf{A}}}}{{d{{\bf{p}}_m}}}{\bf{T}} + {\bf{A}}\frac{{d{\bf{T}}}}{{d{{\bf{p}}_m}}}
=\frac{{d{\bf{A}}}}{{d{{\bf{p}}_m}}}{\bf{T}} + {\bf{A}}\left( {\frac{{d{\bf{T}}}}{{d{\bf{L}}}}} \right)\left( {\frac{{d{\bf{L}}}}{{d{{\bf{p}}_m}}}} \right) \nonumber\\
= &{}\frac{{d{\bf{A}}}}{{d{{\bf{p}}_m}}}{\bf{T}} + {\bf{A}}{{\bf{K}}_c}{{\bf{A}}^T} = :{{\bf{K}}_T} + {{\bf{K}}_k}
\label{eq:J5_3-5}
\end{align}
where ${{\bf{p}}_m}$, ${\bf T}$, and ${\bf L}$ represent the position and orientation of the center of mass of the mobile platform, cable tension vector, and cable position vector, respectively. Matrices ${{\bf{K}}_T}$ and ${{\bf{K}}_k}$ are a product of the cable tensions and cable stiffness, respectively, where ${{\mathbf{K}}_c} = \frac{{d{\mathbf{T}}}}{{d{\mathbf{L}}}} = {\rm{diag}}\left( {{k_1},{k_2}, \cdots ,{k_i}, \cdots ,{k_n}} \right) \in {\mathbb{R}^{n \times n}}$ and ${k_i}$ represents the cable stiffness, i.e., the stiffness coefficient of the $i$th cable. If \eqref{eq:J5_3-5} is expanded in terms of the kinematic parameters ${L_i}$, ${{\mathbf{\hat L}}_i}$, and ${{\mathbf{r}}_i}$, the matrices ${{\mathbf{K}}_T}$ and ${{\mathbf{K}}_k}$ can be described as~\cite{Mendez2014,Behzadipour2006}
\begin{align}
{{\bf{K}}_T} = \sum\limits_{i = 1}^n {\frac{{{T_i}}}{{{L_i}}}\begin{bmatrix}
{{\bf{I}} - {{{\bf{\hat L}}}_i}{\bf{\hat L}}_i^T}&{\left( {{\bf{I}} - {{{\bf{\hat L}}}_i}{\bf{\hat L}}_i^T} \right){{\left[ {{{\bf{r}}_i} \times } \right]}^T}}\\
{\left[ {{{\bf{r}}_i} \times } \right]\left( {{\bf{I}} - {{{\bf{\hat L}}}_i}{\bf{\hat L}}_i^T} \right)}&
\begin{matrix}
\left[ {{{\bf{r}}_i} \times } \right]\left( {{\bf{I}} - {{{\bf{\hat L}}}_i}{\bf{\hat L}}_i^T} \right){{\left[ {{{\bf{r}}_i} \times } \right]}^T} \\
- \left[ {{{{\bf{\hat L}}}_i} \times } \right]{{\left[ {{{\bf{r}}_i} \times } \right]}^T}
\end{matrix}
\end{bmatrix}}
\label{eq:J5_3-6}
\end{align}
and
\begin{align}
{{\bf{K}}_k} = \sum\limits_{i = 1}^n {{k_i}\begin{bmatrix}
{{{{\bf{\hat L}}}_i}{\bf{\hat L}}_i^T}&{{{{\bf{\hat L}}}_i}{\bf{\hat L}}_i^T{{\left[ {{{\bf{r}}_i} \times } \right]}^T}}\\
{\left[ {{{\bf{r}}_i} \times } \right]{{{\bf{\hat L}}}_i}{\bf{\hat L}}_i^T}&{\left[ {{{\bf{r}}_i} \times } \right]{{{\bf{\hat L}}}_i}{\bf{\hat L}}_i^T{{\left[ {{{\bf{r}}_i} \times } \right]}^T}}
\end{bmatrix}}
\label{eq:J5_3-7}
\end{align}
where ${{\bf{r}}_i} = \begin{bmatrix}
{{r_{ix}}}\\
{{r_{iy}}}\\
{{r_{iz}}}
\end{bmatrix}$, $\left[ {{{\bf{r}}_i} \times } \right] = \begin{bmatrix}
0&{ - {r_{iz}}}&{{r_{iy}}}\\
{{r_{iz}}}&0&{ - {r_{ix}}}\\
{ - {r_{iy}}}&{{r_{ix}}}&0
\end{bmatrix}$, ${{\bf{\hat L}}_i} = \begin{bmatrix}
{{{\hat L}_{ix}}}\\
{{{\hat L}_{iy}}}\\
{{{\hat L}_{iz}}}
\end{bmatrix}$, and $\left[ {{{{\bf{\hat L}}}_i} \times } \right] = \begin{bmatrix}
0&{ - {{\hat L}_{iz}}}&{{{\hat L}_{iy}}}\\
{{{\hat L}_{iz}}}&0&{ - {{\hat L}_{ix}}}\\
{ - {{\hat L}_{iy}}}&{{{\hat L}_{ix}}}&0
\end{bmatrix}$. $\left[ {{{\bf{r}}_i} \times } \right]$ is defined as the cross product operator, ${k_i}$ is the $i$th cable stiffness, and $\bf{I}$ is the identity matrix. Eq. \eqref{eq:J5_3-6} and \eqref{eq:J5_3-7} are equivalent to the results of the four-spring model proposed by Behzadipour and Khajepour~\cite{Behzadipour2006}. They also proved that a static cable-driven robot is stable if the stiffness matrix ${\bf{K}}$ is positive definite (sufficient condition).

The stiffness matrices in \eqref{eq:J5_3-6} and \eqref{eq:J5_3-7} are applied to the cable-driven robots in 3D. For the HCDPR in this paper, since the upper four cables are utilized for position control and the lower cables are used to set cable tensions, the specific stiffness matrix can be rearranged as
\begin{align}
{{\bf{K}}_T} = \sum\limits_{i = 1}^6 {\frac{{{T_i}}}{{{L_i}}}} \begin{bmatrix}
{{\bf{I}} - {{{\bf{\hat L}}}_i}{\bf{\hat L}}_i^T}&{\left( {{\bf{I}} - {{{\bf{\hat L}}}_i}{\bf{\hat L}}_i^T} \right){{\left[ {{{\bf{r}}_i} \times } \right]}^T}}\\
{\left[ {{{\bf{r}}_i} \times } \right]\left( {{\bf{I}} - {{{\bf{\hat L}}}_i}{\bf{\hat L}}_i^T} \right)}&
\begin{matrix}
\left[ {{{\bf{r}}_i} \times } \right]\left( {{\bf{I}} - {{{\bf{\hat L}}}_i}{\bf{\hat L}}_i^T} \right){{\left[ {{{\bf{r}}_i} \times } \right]}^T} \\
- \left[ {{{{\bf{\hat L}}}_i} \times } \right]{{\left[ {{{\bf{r}}_i} \times } \right]^T}}
\end{matrix}
\end{bmatrix}
\label{eq:J5_3-8}
\end{align}
and
\begin{align}
{{\bf{K}}_k} = \sum\limits_{i = 1,2,5,6}^n {{k_i}} \begin{bmatrix}
{{{{\bf{\hat L}}}_i}{\bf{\hat L}}_i^T}&{{{{\bf{\hat L}}}_i}{\bf{\hat L}}_i^T{{\left[ {{{\bf{r}}_i} \times } \right]}^T}}\\
{\left[ {{{\bf{r}}_i} \times } \right]{{{\bf{\hat L}}}_i}{\bf{\hat L}}_i^T}&{\left[ {{{\bf{r}}_i} \times } \right]{{{\bf{\hat L}}}_i}{\bf{\hat L}}_i^T{{\left[ {{{\bf{r}}_i} \times } \right]}^T}}
\end{bmatrix}
\label{eq:J5_3-9}
\end{align}
where ${r_{iy}}$ and ${\hat L_{iy}}$ equal zero, and ${k_i} = {K_s}/{L_{0i}}$. In addition, elements of ${{\bf{K}}_k}$ cannot be controlled, because they come from the property of the cables. Hence, the stiffness of HCDPR can be changed by optimizing ${{\bf{K}}_T}$.

Then, maximizing the stiffness of HCDPR is achieved by the following approach:

1) When the kinematic constraints (${L_{01}} = {L_{02}}$ and ${L_{05}} = {L_{06}}$) are applied, then set the two cable tensions as ${T_1} = {T_2}$ and ${T_5} = {T_6}$. By combining ${k_i} = {K_s}/{L_{0i}}$ and \eqref{eq:J5_26}, we get
\begin{align}
\left\{ \begin{array}{l}
{\lambda _1} = \frac{{{b_2}{c_1} - {b_1}{c_2}}}{{{a_1}{b_2} - {a_2}{b_1}}}\\
{\lambda _2} = \frac{{{a_1}{c_2} - {a_2}{c_1}}}{{{a_1}{b_2} - {a_2}{b_1}}}
\end{array} \right.
\label{eq:J5_3-10}
\end{align}
where
\begin{align}
\left\{ \begin{array}{l}
{a_1} = {N_{A11}} - {N_{A21}}\\
{b_1} = {N_{A12}} - {N_{A22}}\\
{c_1} = {k_1}\left( {{L_1} - {L_2}} \right) + {T_{A2}} - {T_{A1}} + \left( {{N_{A23}} - {N_{A13}}} \right){\lambda _3}\\
{a_2} = {N_{A51}} - {N_{A61}}\\
{b_2} = {N_{A52}} - {N_{A62}}\\
{c_2} = {k_5}\left( {{L_5} - {L_6}} \right) + {T_{A6}} - {T_{A5}} + \left( {{N_{A63}} - {N_{A53}}} \right){\lambda _3}.
\end{array} \right.
\end{align}

Hence, there is only one variable ${\lambda _3}$ to optimize, such that:
\begin{align}
{\bf{T}} = {{\bf{T}}_A} + {{\bf{N}}_A}{\begin{bmatrix}
{\frac{{{b_2}{c_1} - {b_1}{c_2}}}{{{a_1}{b_2} - {a_2}{b_1}}}}&{\frac{{{a_1}{c_2} - {a_2}{c_1}}}{{{a_1}{b_2} - {a_2}{b_1}}}}&{{\lambda _3}}
\end{bmatrix}^T}.
\label{eq:J5_3-11}
\end{align}

2) Maximizing any diametric matrix $\bf{K}$ in \eqref{eq:J5_3-5} provides a unique solution for cable tensions~\cite{G.Meunier2009} and satisfies ${\bf{K}} = g({\lambda _3})$, where $g( \cdot )$ is a monotonic nondecreasing function. So, maximizing the stiffness $\bf{K}$ and maximizing ${\lambda _3}$ are equivalent. Here, the maximum ${\lambda _3}$ will maximize the HCDPR's stiffness $\bf{K}$ in ${\color{black}x_m}$, ${\color{black}z_m}$, and ${\color{black}\theta_m}$ directions (i.e., in the directions of X-axis, Z-axis, and rotation about Y-axis). Then, we have
\begin{align}
{\mathbf{T}} = {\lambda _3}{{\mathbf{D}}_A} + {{\mathbf{E}}_A},\quad \quad {{\mathbf{D}}_A},{{\mathbf{E}}_A} \in {\mathbb{R}^6}
\label{eq:J5_3-12}
\end{align}
where matrices ${{\mathbf{D}}_A}$ and ${{\mathbf{E}}_A}$ are calculated using \eqref{eq:J5_appendix_B_1} and \eqref{eq:J5_appendix_B_2} shown in Appendix~\ref{appendix:J5_SomeTermsStiffness}. The solution for \eqref{eq:J5_3-12}  can be described as
\begin{align}
{\lambda _3} = \frac{1}{{{D_{Ai}}}}{T_i} - \frac{{{E_{Ai}}}}{{{D_{Ai}}}},\quad \quad i = 1,2, \cdots ,6.
\label{eq:J5_3-13}
\end{align}

3) The objective function is defined as
\begin{subequations} 
\begin{align}
{{\rm{maximize}}}\quad&{{\lambda _3}} \\
{\rm{subject{~}to}}\quad&{\bf{T}} = {\lambda _3}{{\bf{D}}_A} + {{\bf{E}}_A}\\
&{0 \le {T_{i\min }} \le {T_i} \le {T_{i\max }},\quad \quad i = 1,2, \cdots ,6}
\end{align}
\label{eq:J5_3-14}
\end{subequations}
where ${T_i},\;{T_{i\min }},$ and ${T_{i\max }}$ represent the $i$th cable tension, minimum allowable tension, and maximum allowable tension, respectively. Eq.~\eqref{eq:J5_3-14} can be easily solved using solvers such as CVX~\cite{cvx} to find the optimal value ${\lambda _3}$. After ${\lambda _3}$ is obtained, the corresponding optimal cable tension ${\bf{T}}$ is calculated using \eqref{eq:J5_3-12}. Compared to the method of stiffness maximization in the softest direction in~\cite{H.Jamshidifar2017}, \eqref{eq:J5_3-14} provides a simpler and effective approach. In this research, the above algorithm (used to calculate ${\bf{T}}$ from \eqref{eq:J5_3-12}) is combined with controller design to meet the control objective while simultaneously satisfying required stiffness along each motion axes.

%
\subsection{Control Strategies}\label{subsec:J5_ControlStrategies}
For the proposed configuration of the HCDPR shown in~\autoref{fig:J5_PlanarModel}, four upper cables, two lower cables, and the 2-DOF robot arm are based on position control, force control, and torque control, respectively, i.e., their corresponding inputs are positions (cable lengths), forces, and joint torques. Furthermore, the elastic cables reduce the overall stiffness of the robot, so vibrations become a serious problem for precise control~\cite{Behzadipour2006,G.Meunier2009,M.Rushton2016Acc}. Another major problem is maintaining cable tensions to keeping large enough stiffness for the robot. As mentioned above, the goal of this paper is to develop an integrated control system for the HCDPR to reduce vibrations and improve motion accuracy and performance. In order to achieve this goal, different controllers can be designed, such as PID, LQR, and feed-forward controllers. In the studies, PID-based control strategies are designed to control the motion of the HCDPR.

In addition, for the actual HCDPR system, since the driven cables are flexible, the positions of the mobile platform or actual cable lengths cannot be computed directly from the measurements of encoders (embedded in the corresponding driven actuators). In this case, the upper cables are considered as a rigid body with the given cable lengths while optimizing the lower cable tensions. In other words, each cable is fitted with a force sensor which provides tension magnitude to the robot feedback control system to ensure that the HCDPR has the desired stiffness. The control strategy includes tracking the motions of the mobile platform and the robot arm as well as optimizing the cable tensions to satisfy the required stiffness of the robot.

\begin{figure*}[!t]\centering
	\includegraphics[width=130mm]{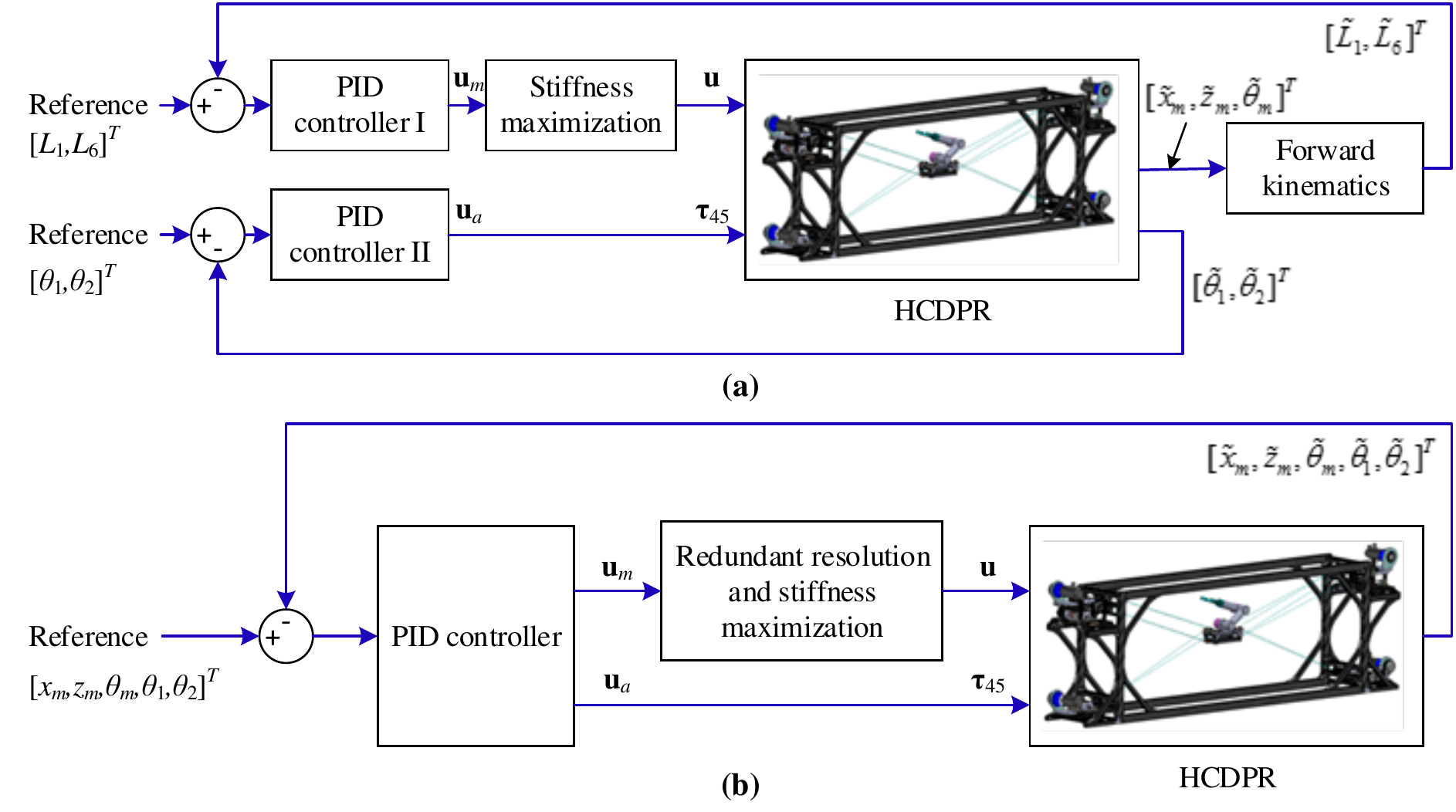}
	\caption{Control structures of the HCDPR. (a) The desired inputs are cable lengths and robot arm joint variables; (b) The desired inputs are joint variables.}\label{fig:J5_ControlStructure}
\end{figure*}

Based on the above method, the proposed control structures of the HCDPR are shown in~\autoref{fig:J5_ControlStructure}. \autoref{fig:J5_ControlStructure}(a) shows the desired inputs being cable lengths $({\color{black}{L_{1}},{L_{6}}})$ and robot arm joint variables $({\color{black}\theta_{1}},{\color{black}\theta_{2}})$. In this case, the goal is to control the rigid HCDPR for the desired $({\color{black}{L_{1}},{L_{6}}})$ using PID controller I and the desired $({\color{black}\theta_{1}},{\color{black}\theta_{2}})$ using PID controller II, respectively. \autoref{fig:J5_ControlStructure}(b) represents the desired inputs as joint variables $\mathbf{q}=({\color{black}x_m},{\color{black}z_m},{{\color{black}\theta_m}},{\color{black}\theta_{1}},{\color{black}\theta_{2}})$. In this case, suppose $({\color{black}x_m},{\color{black}z_m},{{\color{black}\theta_m}})$ (i.e., ${P_m}({x_m},{z_m},{\color{black}\theta_m})$) is given (e.g., using external cameras to track the trajectories). In the control scheme, the corresponding PID controllers continuously calculate errors as the difference between the desired and actual values. The controllers’ outputs are then used to command the cables and the robot arm actuators to drive the HCDPR. Also, the optimal cable tensions are obtained using \eqref{eq:J5_3-12}.

The HCDPR system with position controllers developed is shown in \autoref{fig:J5_ControlStructure}(a) and \autoref{fig:J5_ControlStructure}(b). The system consists of the system dynamics in~\autoref{sec:J5_SystemModeling}, the redundancy resolution derived in~\autoref{subsec:J5_RedundancyResolution}, and the stiffness maximization approach proposed in~\autoref{subsec:J5_MaxStiffness}.

Defining an error vector $\mathbf{e}(t)$ for the controllers above as
\begin{align}
\mathbf{e}(t) = 
\left\{\begin{array}{l}
[L_1(t) \;\; L_6(t)]^T- [{\tilde L_1}(t)\;\; {\tilde L_6}(t)]^T \;\;\;\: {\rm{PID{~}controller{~}I}}\\
[\theta_1(t) \;\; \theta_2(t)]^T- [{\tilde \theta_1}(t)\;\; {\tilde \theta_2}(t)]^T \quad {\rm{PID{~}controller{~}II}}\\
\mathbf{q}(t) - \tilde {\mathbf{q}}(t)\\
\end{array} \right.
\label{eq:J5_3-15}
\end{align}
where $\tilde{( \cdot )}$ denotes actual values. Based on the diagram shown in~\autoref{fig:J5_ControlStructure}, the control law is designed as
\begin{align}
\begin{bmatrix}
\mathbf{u}_m(t)\\
\mathbf{u}_a(t)
\end{bmatrix}
 = {K_p}\mathbf{e}(t) + {K_i}\int_0^t {\mathbf{e}(t)dt}  + {K_d}\frac{{d\mathbf{e}(t)}}{{dt}}
\label{eq:J5_3-16}
\end{align}
where ${K_p}$, ${K_i}$, and ${K_d}$ are the proportional, integral, and derivative terms, respectively. $\mathbf{u}_m$ and $\mathbf{u}_a$ represent control inputs to the mobile platform and robot arm, respectively.

\section{Numerical Results and Discussion}
\label{sec:J5_NumericalResults}
To evaluate the control performance in~\autoref{sec:J5_ControlDesign}, the following cases will be studied. All the scenarios are implemented using MATLAB 2019a (The MathWorks, Inc.) on a Windows 7 x64 desktop PC (Inter Core i7-4770, 3.4 GHz CPU and 8 GB RAM), and the initial condition is ${{\bf{q}}_0} = {[{\rm{0}},{\rm{0}},0,0,0]^T}$.

\begin{itemize}
\item Case 1: Control the CDPR by Given Cable Lengths $({\color{black}{L_{1}},{L_{6}}})$
\end{itemize}\par
\begin{figure}[h]
\centering
\subfigure[]{\label{subfig:J5_Case_1_a}\includegraphics[width=42mm]{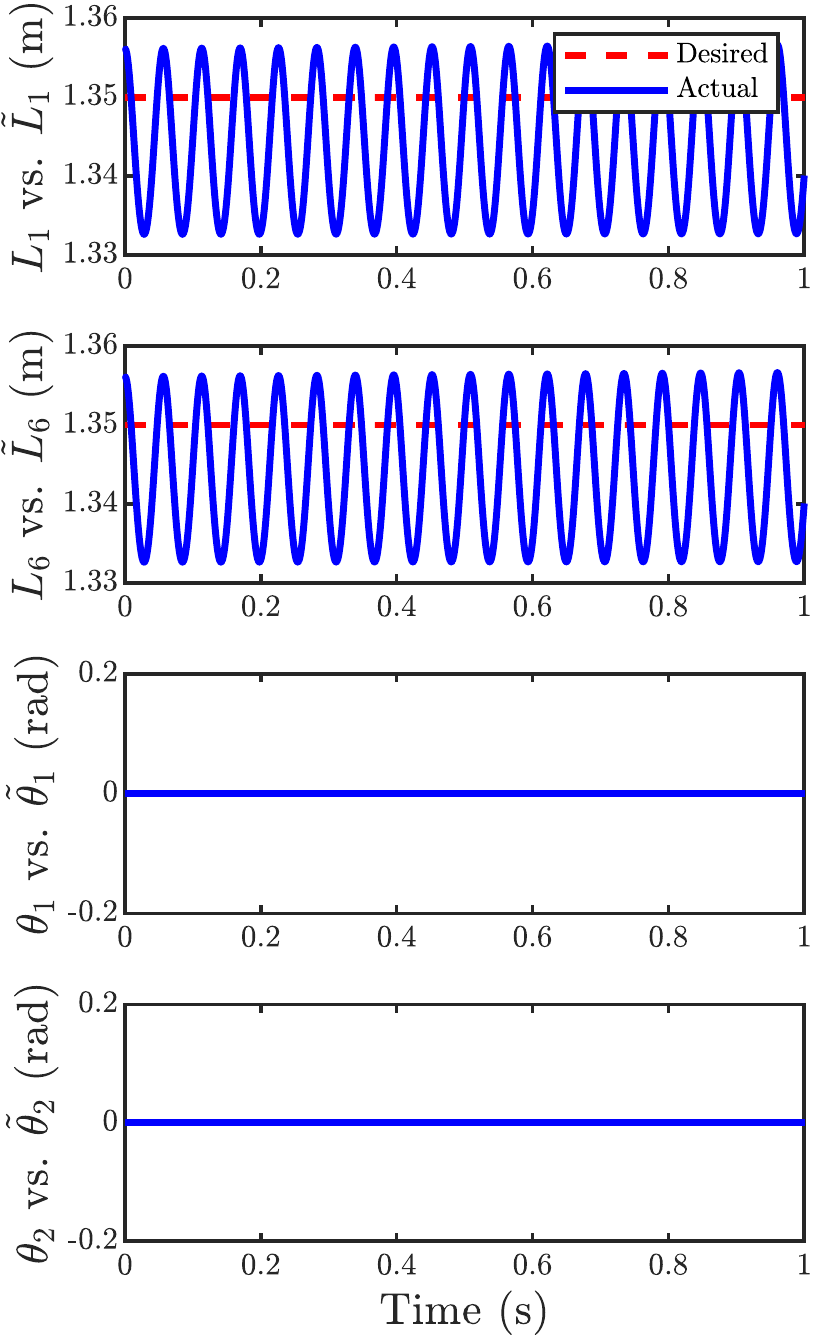}}
\subfigure[]{\label{subfig:J5_Case_1_b}\includegraphics[width=42mm]{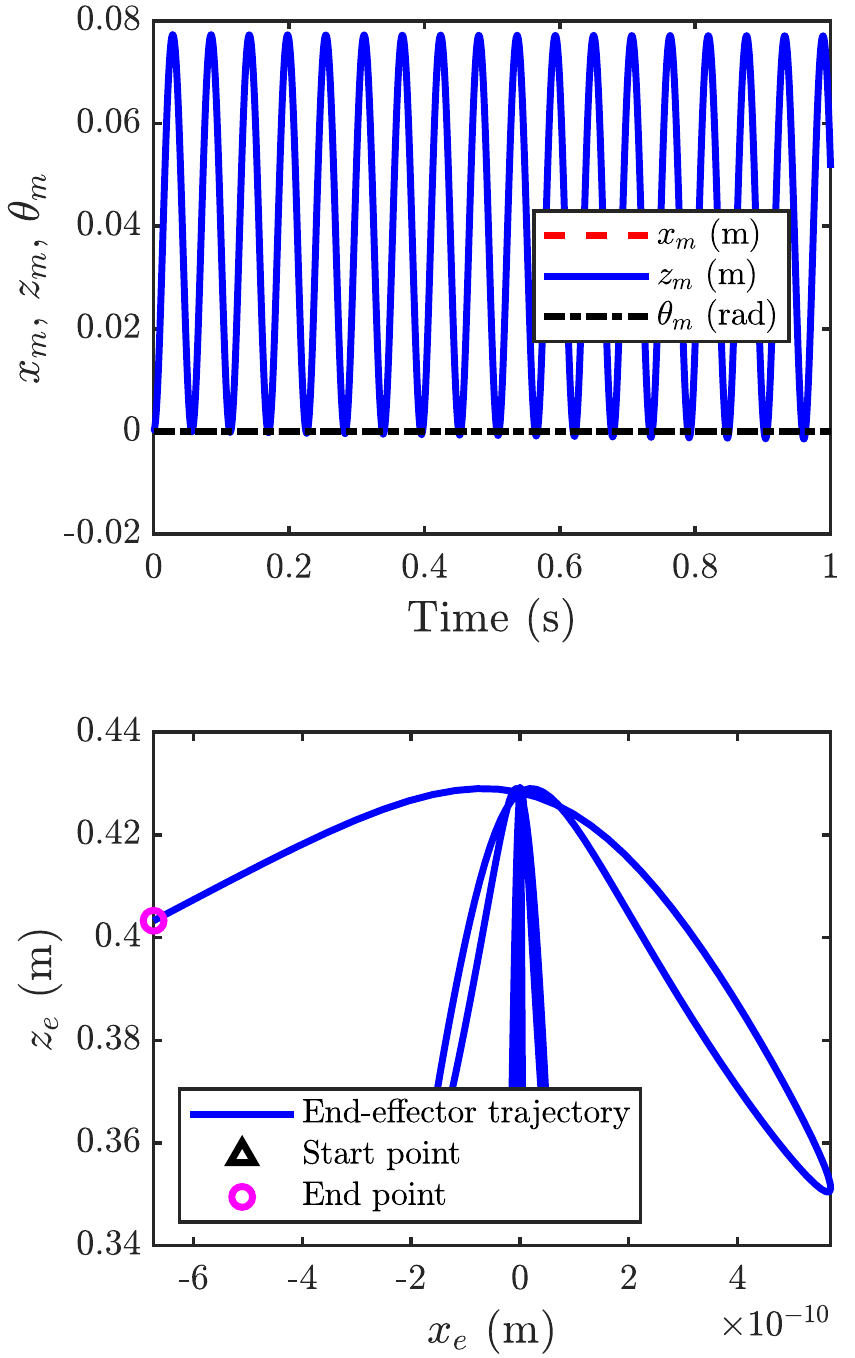}}\\
\caption{Responses of Case 1. (a) Errors between the desired inputs and the actual outputs and (b) trajectories of the center of mass of the mobile platform and the end-effector.}
\label{fig:J5_Case_1}
\end{figure}

In the first case, assume ${\color{black}{L_{1}},{L_{6}}},{\color{black}\theta_{1}}$, and ${\color{black}\theta_{2}}$ are obtained from the actuators encoders. The PID controller I is applied (parameters are set as $K_p = 2 \times 10^2, K_i = 10, {~}{\rm{and}}{~} K_d = 0$) to control the upper cables, where ${{\color{black}L_{1}}} = 1.35~\rm{m}$ and ${{\color{black}L_{6}}} = 1.35~\rm{m}$. Meanwhile, the PID controller II is applied (parameters are set as $K_p = 6 \times 10^2, K_i = 20, {~}{\rm{and}}{~} K_d = 1 \times 10^2$) to control the robot arm, and the desired joint variables $[{\color{black}\theta_{1}},{\color{black}\theta_{2}}]$ are equal to $[0,0]$. This means that it is desired to maintain the robot arm stationary. The results in Figure~\ref{subfig:J5_Case_1_a} and Figure~\ref{subfig:J5_Case_1_b} show that the vibrations are not damped out using the proposed PID controllers.

\begin{itemize}
\item Case 2: Control the CDPR by Given ${P_m}({x_m},{z_m},{\color{black}\theta_m})$
\end{itemize}\par
\begin{figure}[!t]
\centering
\subfigure[]{\label{subfig:J5_Case_2_a}\includegraphics[width=42mm]{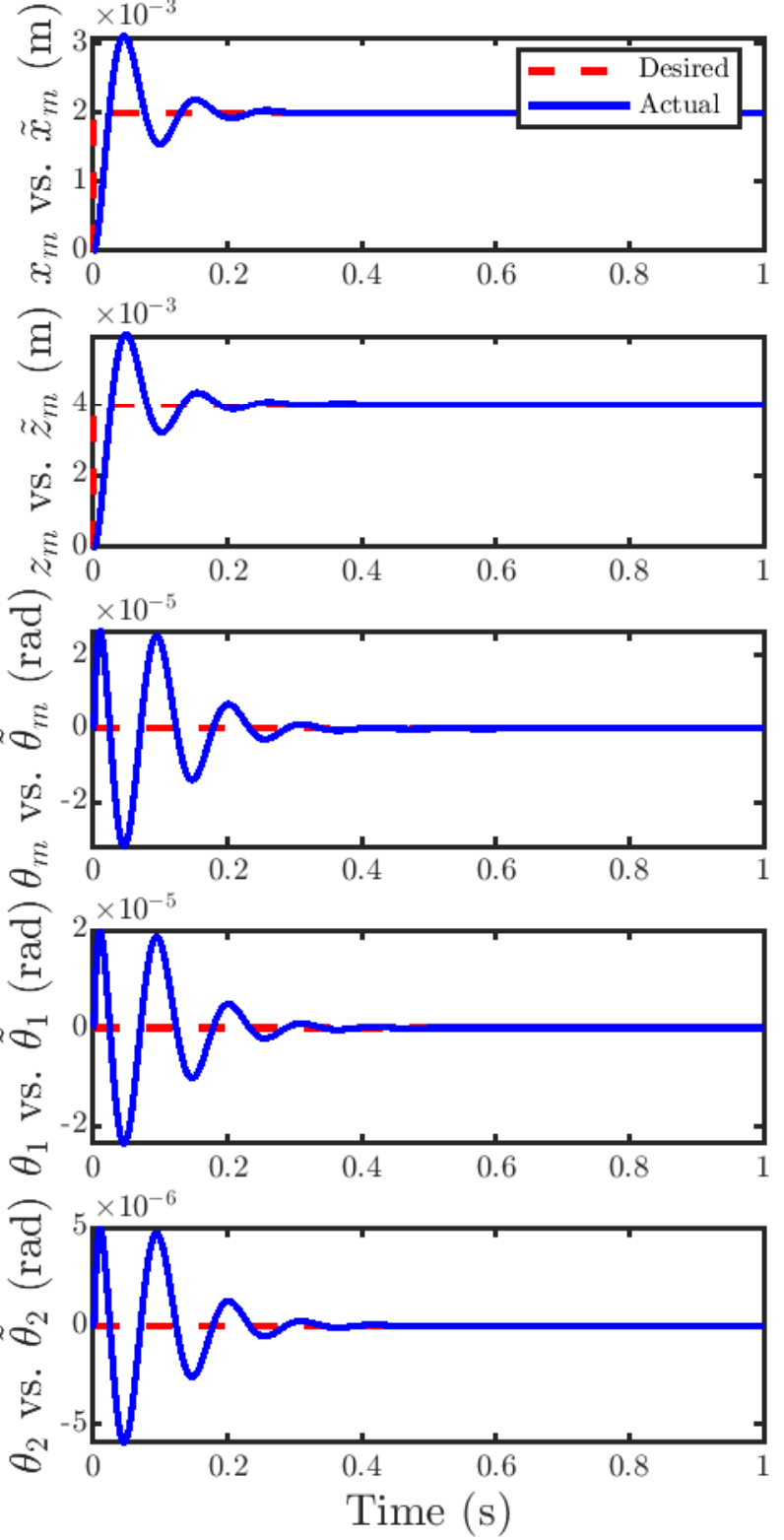}}
\subfigure[]{\label{subfig:J5_Case_2_b}\includegraphics[width=42mm]{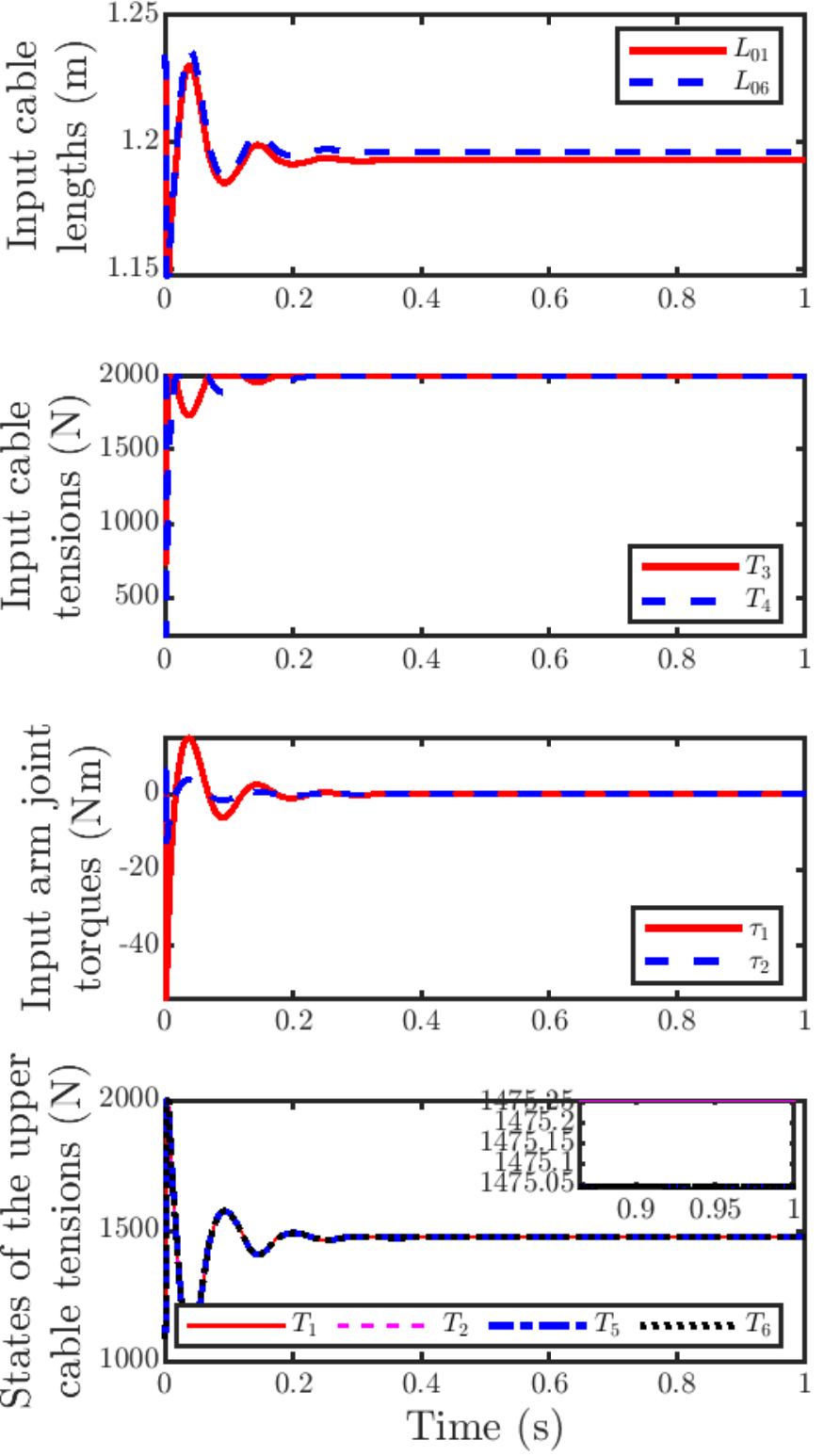}}\\
\subfigure[]{\label{subfig:J5_Case_2_c}\includegraphics[width=84mm]{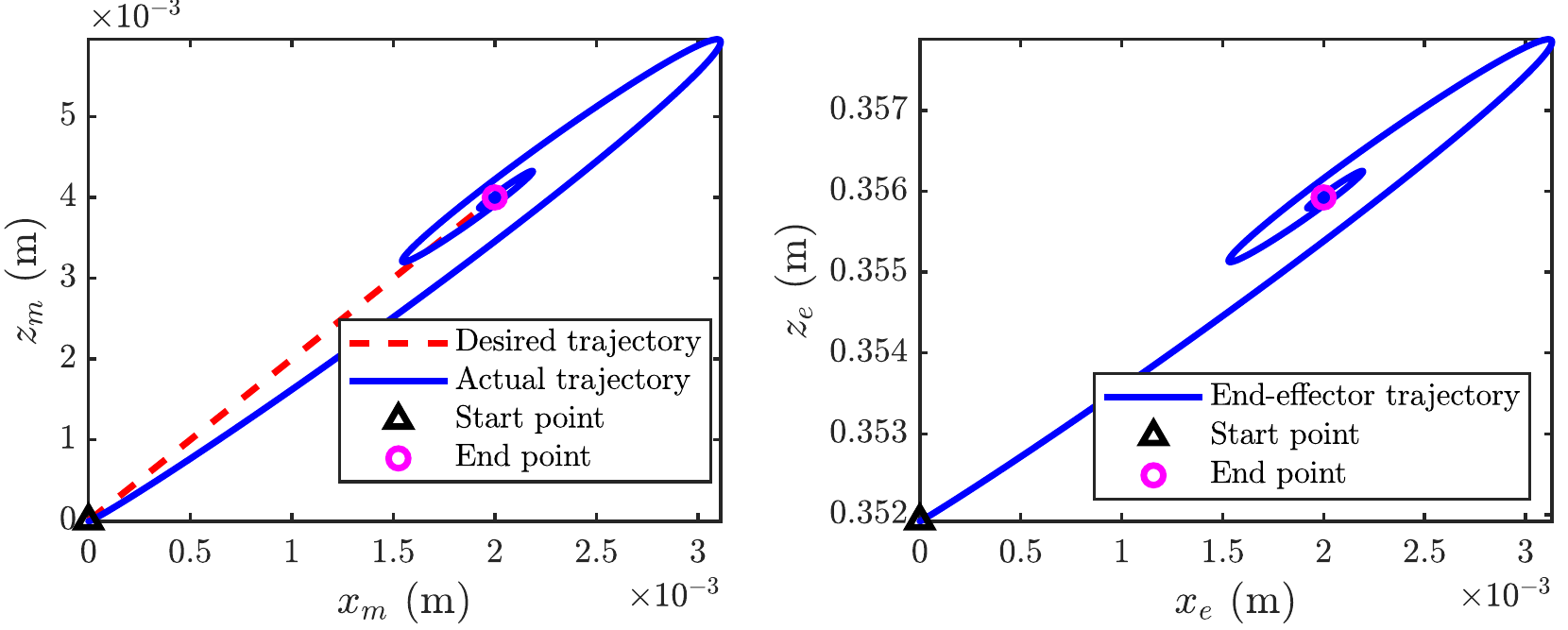}}\\
\caption{Responses of Case 2. (a) Errors between the desired input ${\bf{q}}$ and the actual output ${\bf{\tilde q}}$, (b) the dynamic inputs (cable lengths, cable tensions, and robot arm joint torques) applied to the HCDPR and the states of the upper cable tensions, and (c) trajectories of the center of mass of the mobile platform and the robot arm end-effector.}
\label{fig:J5_Case_2}
\end{figure}

In this case, suppose ${P_m}({x_m},{z_m},{\color{black}\theta_m})$ (i.e., $({\color{black}x_m},{\color{black}z_m},{{\color{black}\theta_m}})$ measurements are available (e.g., vision based feedback). When the PID controller is applied ($K_p = 5 \times 10^5, K_i = 3.5 \times 10^7, {~}{\rm{and}}{~} K_d = 1.1 \times 10^4$), the desired positions of the mobile platform $[{\color{black}x_m},{\color{black}z_m},{{\color{black}\theta_m}},{\color{black}\theta_{1}},{\color{black}\theta_{2}}]$ are set to $[{{2 \times 10^{-3}}},{{4 \times 10^{-3}}},0,0,0]$. The corresponding results are shown in Figure~\ref{subfig:J5_Case_2_a}, Figure~\ref{subfig:J5_Case_2_b}, and Figure~\ref{subfig:J5_Case_2_c}. The results show that the errors between the desired input ${\bf{q}}$ and the actual output ${\bf{\tilde q}}$ go to zero very quickly (about 0.3 seconds), and the dynamic inputs (cable lengths, cable tensions, and robot arm joint torques) applied to the HCDPR are quick to stabilize. In this case, the states of the upper cable tensions stabilize at the set point in less than 0.3 seconds. In addition, it is clear that the vibrations are well controlled when the PID controller is applied.

In summary, based on the results from case 1 and case 2, when the desired ${{\color{black}L_{1}}}$, ${{\color{black}L_{6}}}$, ${\color{black}\theta_{1}}$, and ${\color{black}\theta_{2}}$ are given, vibrations in actual positions of all degrees of freedom need to be damped out (or controlled better).

\begin{itemize}
\item Case 3(a): The Mobile Platform is Fixed and the Robot Arm is Moving
\end{itemize}\par
In this case, the mobile platform is fixed (the cable lengths $({\color{black}{L_{1}},{L_{6}}})$ are given) and the robot arm is moving. Also, PID controller I ($K_p = 2 \times 10^2, K_i = 10, {~}{\rm{and}}{~} K_d = 0$) is applied to control the upper cables and PID controller II ($K_p = 6 \times 10^2, K_i = 20, {~}{\rm{and}}{~} K_d = 1 \times 10^2$) is applied to the robot arm. The desired trajectories are defined as
\begin{align}
\left\{ \begin{array}{l}
{{\color{black}L_{1}}} = {{\color{black}L_{6}}} = 1.35\\
{\color{black}\theta_{1}} = 0.1t,\quad t \in \left[ {0,{t_{\max }}} \right]\\
{\color{black}\theta_{2}} = 0.1t,\quad t \in \left[ {0,{t_{\max }}} \right]
\end{array} \right.
\label{eq:J5_4-1}
\end{align}
where $t$ and ${t_{\max }}$ are the current and maximum running time. 

\begin{figure}[!t]
\centering
\subfigure[]{\label{subfig:J5_Case_3a_a}\includegraphics[width=42mm]{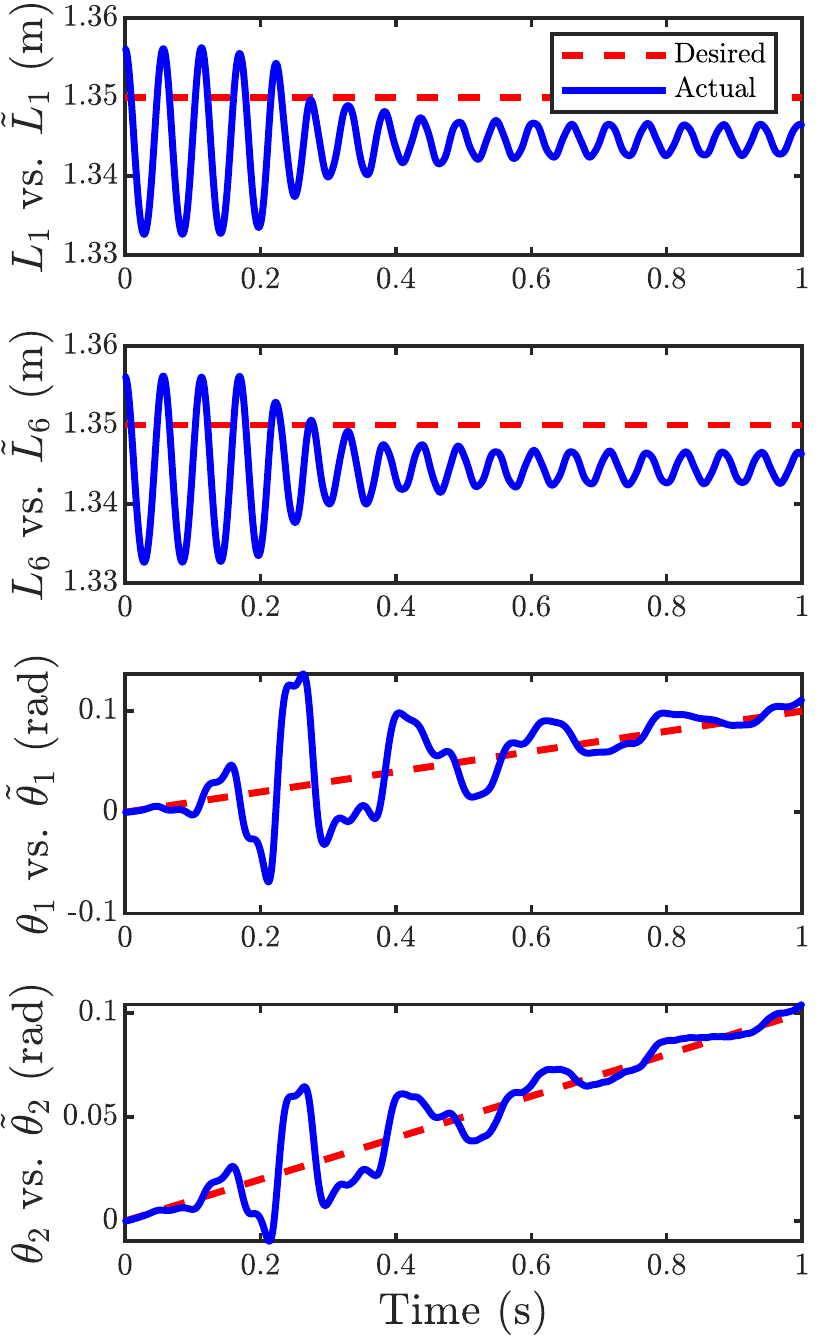}}
\subfigure[]{\label{subfig:J5_Case_3a_b}\includegraphics[width=42mm]{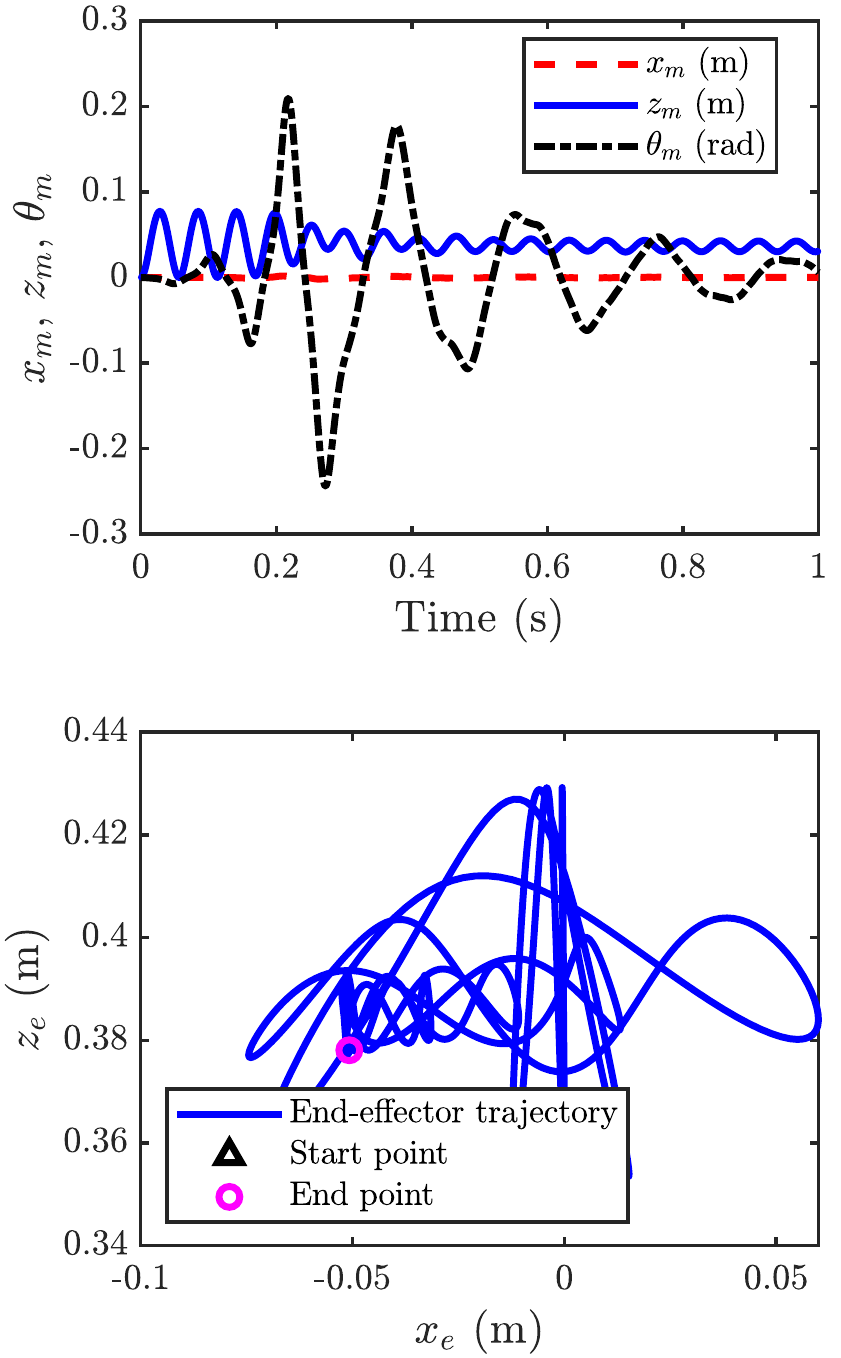}}\\
\caption{Responses of Case 3(a). (a) Errors between the desired inputs and the actual outputs and (b) trajectories of the center of mass of the mobile platform and the end-effector.}
\label{fig:J5_Case_3a}
\end{figure}

The corresponding results are shown in Figure~\ref{subfig:J5_Case_3a_a} and Figure~\ref{subfig:J5_Case_3a_b}. The results also show that tracking errors are not acceptable, and vibrations are not controlled with near sustained oscillations in cables $L_1$ and $L_6$.

\begin{itemize}
\item Case 3(b): The Robot Arm is Fixed and the Mobile Platform is Moving
\end{itemize}\par
In this case, the robot arm is fixed and the mobile platform is moving. This is the same as in Case 3(a), PID controller I ($K_p = 2 \times 10^2, K_i = 10, {~}{\rm{and}}{~} K_d = 0$) is applied to control the upper cables and PID controller II ($K_p = 6 \times 10^2, K_i = 20, {~}{\rm{and}}{~} K_d = 1 \times 10^2$) is applied to the robot arm. The desired trajectories are given by
\begin{align}
\left\{ \begin{array}{l}
{{\color{black}L_{1}}} = 1.35 - 0.01t,\quad t \in \left[ {0,{t_{\max }}} \right]\\
{{\color{black}L_{6}}} = 1.35 + 0.01t,\quad t \in \left[ {0,{t_{\max }}} \right]\\
{\color{black}\theta_{1}} = 0\\
{\color{black}\theta_{2}} = 0
\end{array} \right.
\label{eq:J5_4-2}
\end{align}
where $t$ and ${t_{\max }}$ are the current and maximum running time. 

\begin{figure}[!t]
\centering
\subfigure[]{\label{subfig:J5_Case_3b_a}\includegraphics[width=42mm]{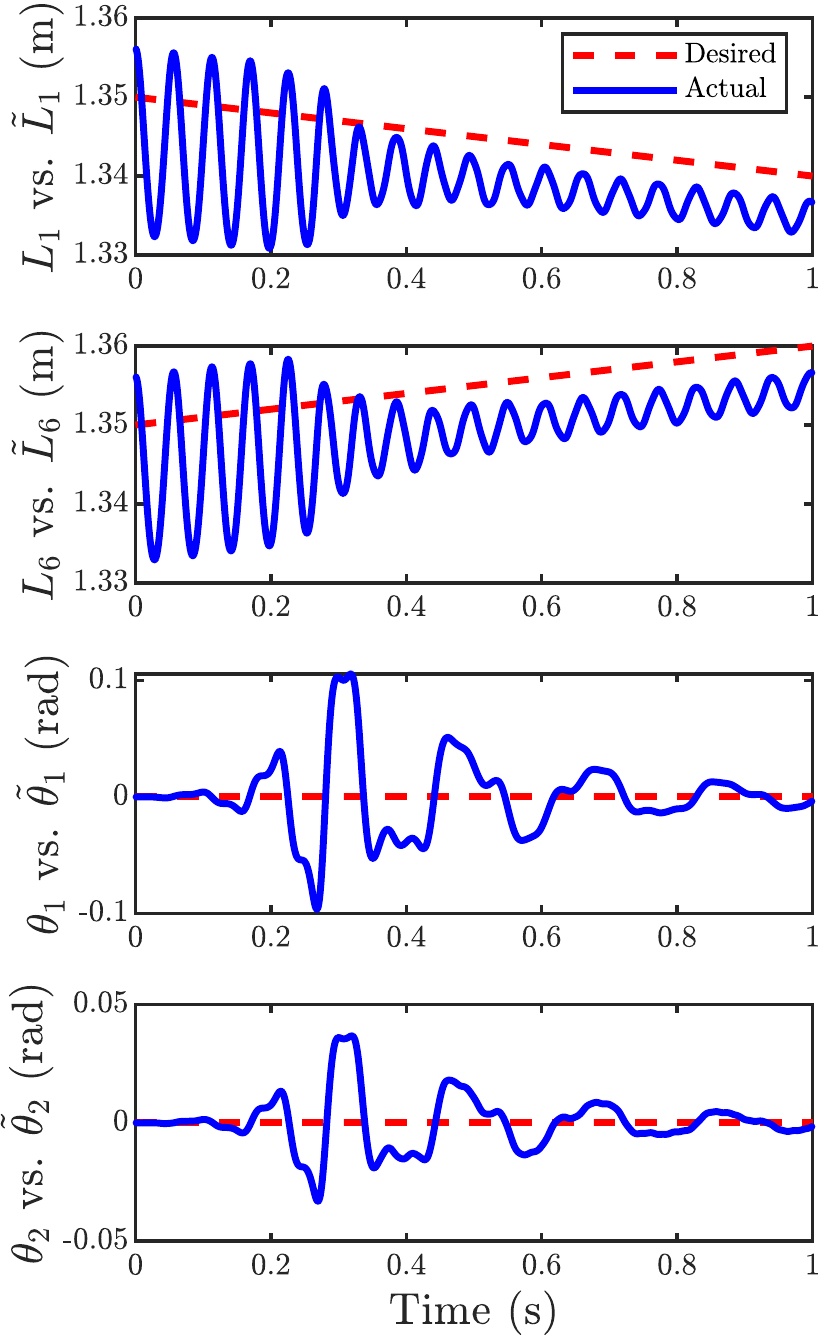}}
\subfigure[]{\label{subfig:J5_Case_3b_b}\includegraphics[width=42mm]{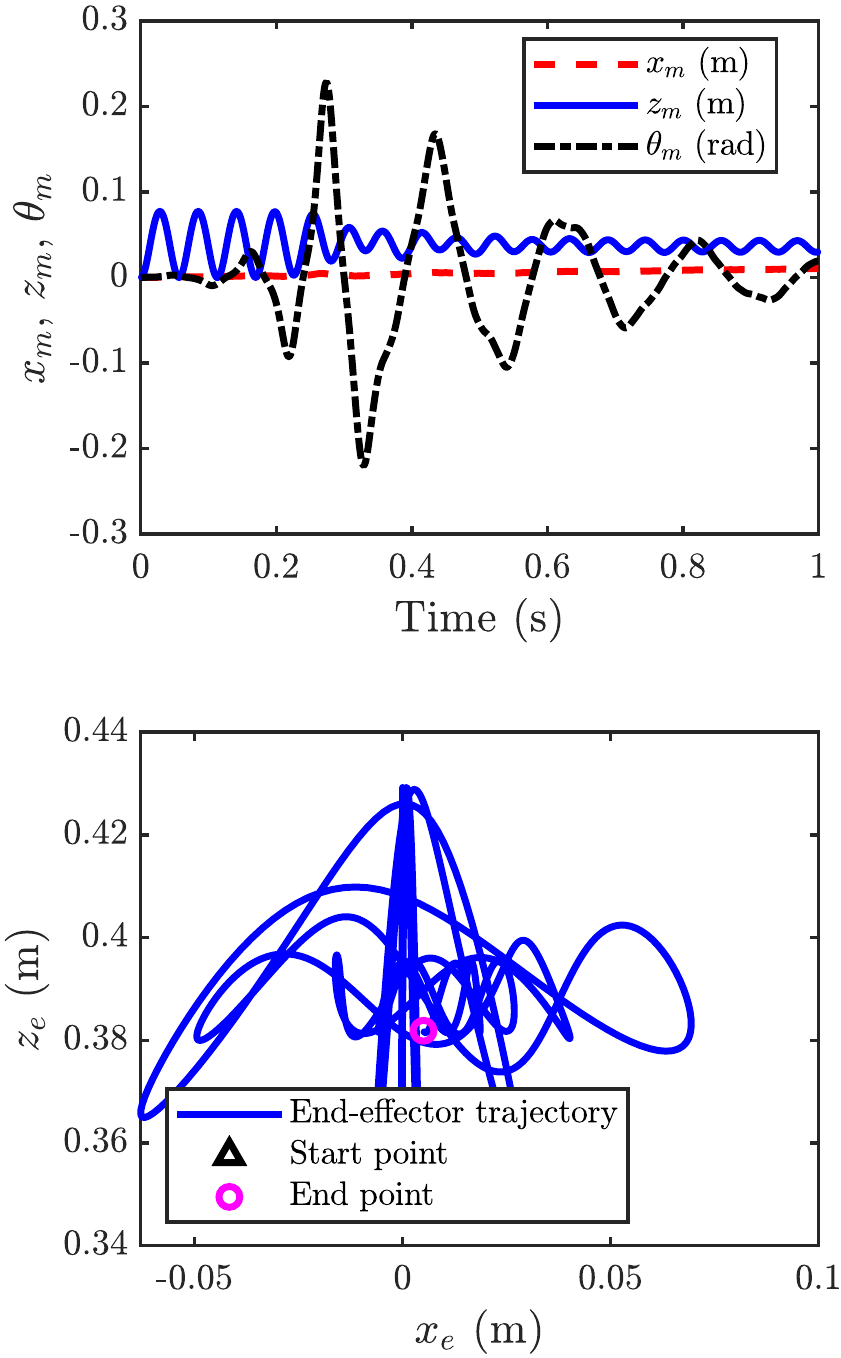}}\\
\caption{Responses of Case 3(b). (a) Errors between the desired inputs and the actual outputs and (b) trajectories of the center of mass of the mobile platform and the end-effector.}
\label{fig:J5_Case_3b}
\end{figure}

In this case, the results are shown in Figure~\ref{subfig:J5_Case_3b_a} and Figure~\ref{subfig:J5_Case_3b_b}. The results again show that tracking errors are not satisfactory and vibrations are not damped out using the two PID controllers.

\begin{itemize}
\item Case 4(a): The Mobile Platform is Fixed and the Robot Arm is Moving
\end{itemize}\par
In this case, the mobile platform is fixed and the robot arm is moving, i.e., the robot arm moves from one point to another. When the PID controller is applied ($K_p = 5 \times 10^5, K_i = 3.5 \times 10^7, {~}{\rm{and}}{~} K_d = 1.1 \times 10^4$), the desired trajectories of the mobile platform are described as
\begin{align}
\left\{ \begin{array}{l}
{\color{black}x_m} = 0\\
{\color{black}z_m} = 0\\
{{\color{black}\theta_m}} = 0\\
{\color{black}\theta_{1}} = t,\quad t \in \left[ {0,{t_{\max }}} \right]\\
{\color{black}\theta_{2}} =  - t,\quad t \in \left[ {0,{t_{\max }}} \right]
\end{array} \right.
\label{eq:J5_4-3}
\end{align}
where $t$ and ${t_{\max }}$ are the current and maximum running time.

\begin{figure}[!t]
\centering
\subfigure[]{\label{subfig:J5_Case_4a_a}\includegraphics[width=42mm]{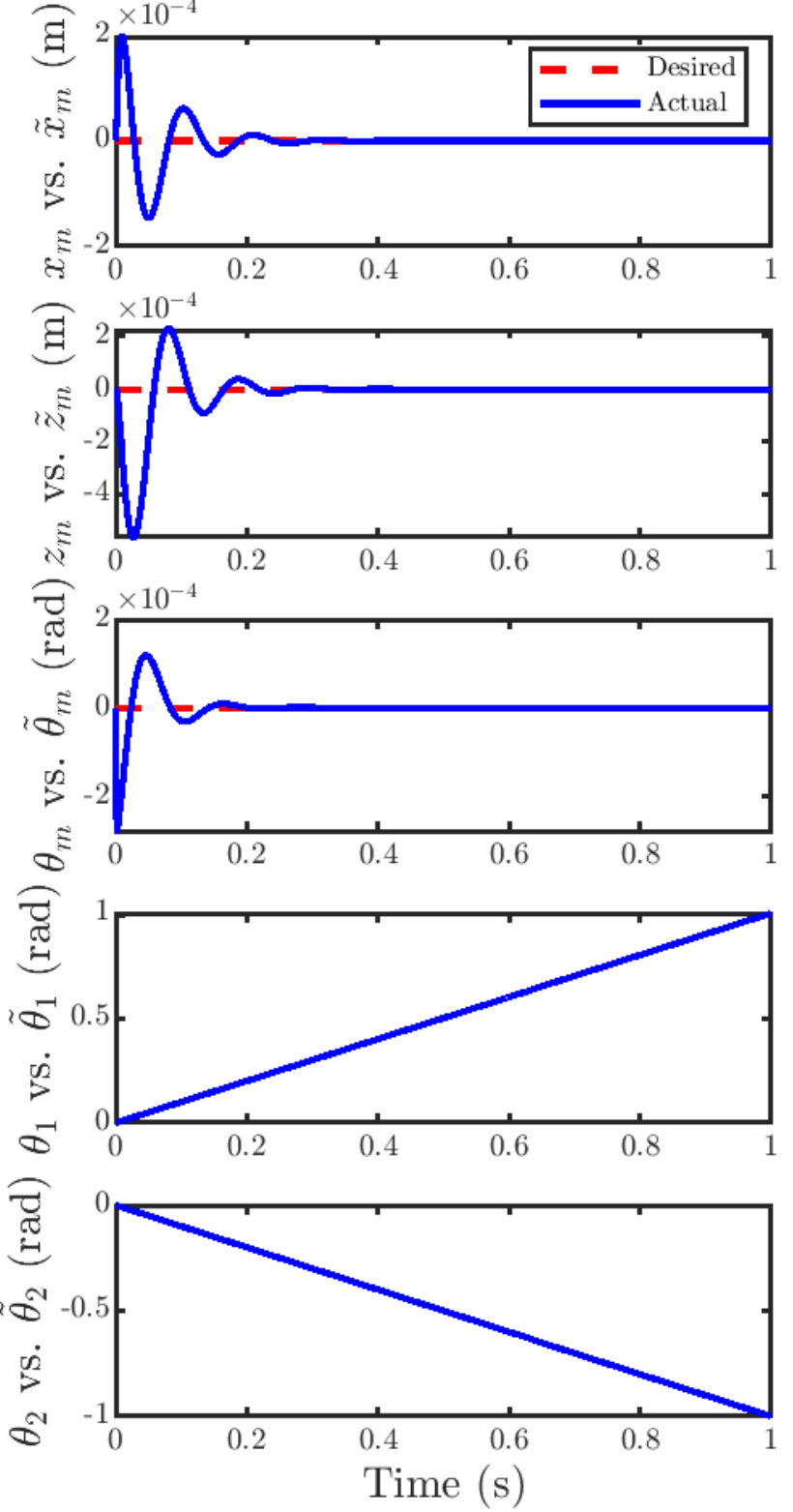}}
\subfigure[]{\label{subfig:J5_Case_4a_b}\includegraphics[width=42mm]{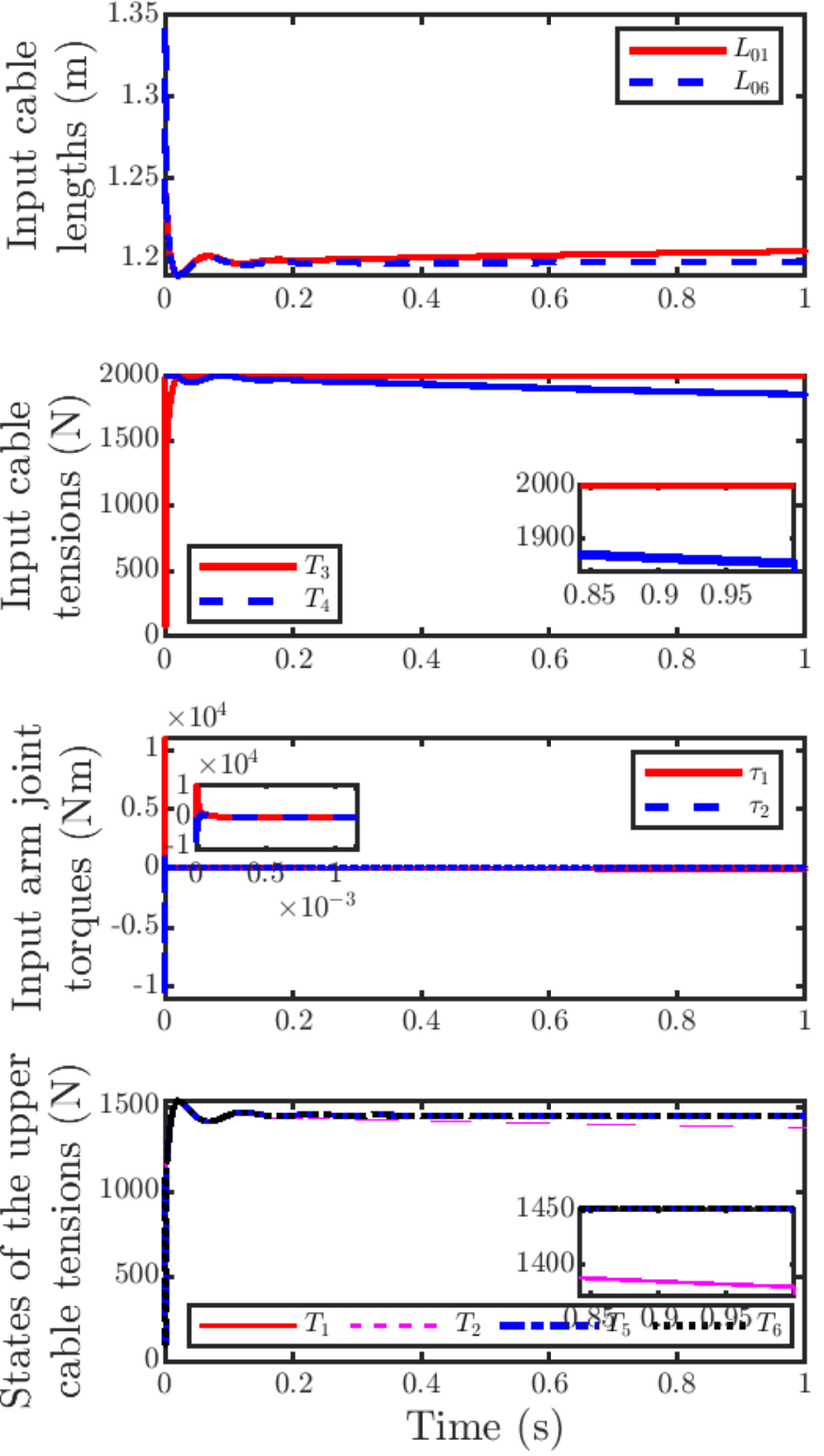}}\\
\subfigure[]{\label{subfig:J5_Case_4a_c}\includegraphics[width=84mm]{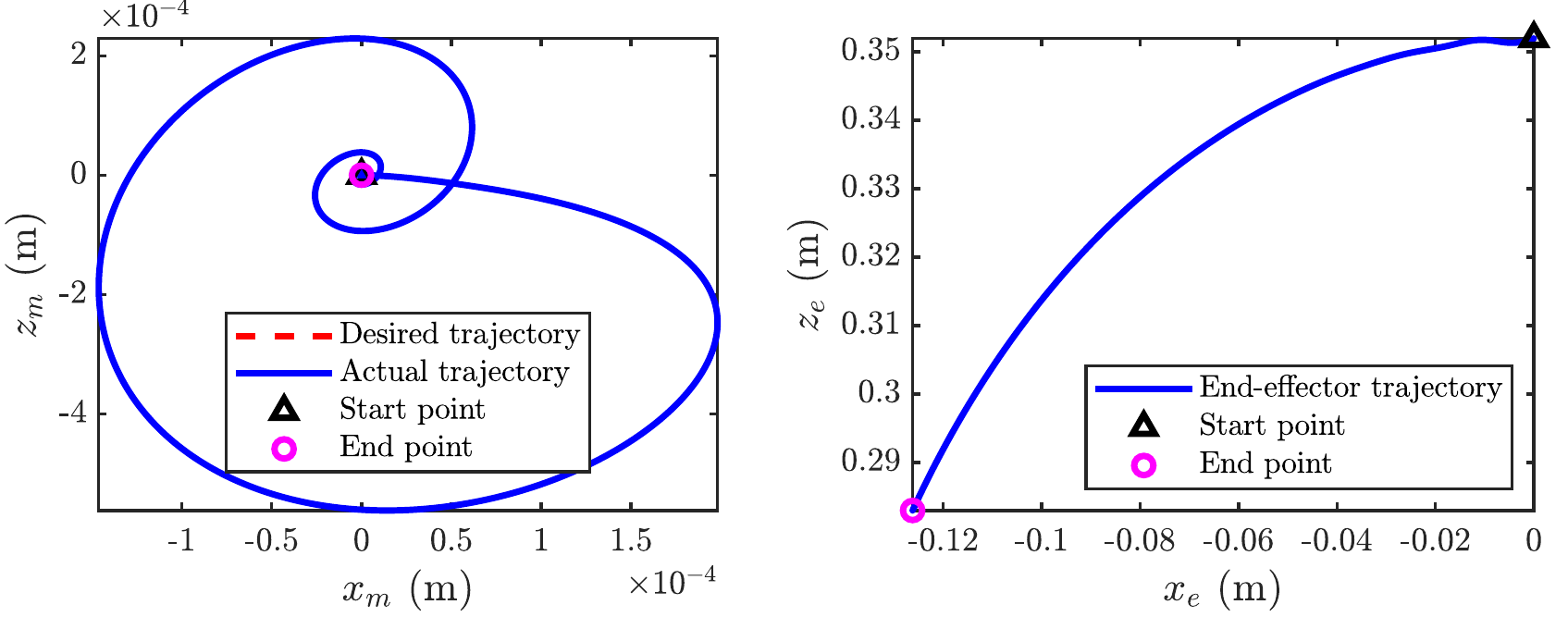}}\\
\caption{Responses of Case 4(a). (a) Errors between the desired input ${\bf{q}}$ and the actual output ${\bf{\tilde q}}$, (b) the dynamic inputs (cable lengths, cable tensions, and arm joint torques) applied to the HCDPR and the states of the upper cable tensions, and (c) trajectories of the center of mass of the mobile platform and the end-effector.}
\label{fig:J5_Case_4a}
\end{figure}

The corresponding results are shown in Figure~\ref{subfig:J5_Case_4a_a}, Figure~\ref{subfig:J5_Case_4a_b}, and Figure~\ref{subfig:J5_Case_4a_c}. The results show that the errors between the desired input ${\bf{q}}$ and the actual output ${\bf{\tilde q}}$ go to zero very quickly, and the dynamic inputs applied to the HCDPR are quick to stabilize. Moreover, although the mobile platform remains stationary and only the robot arm moves from one point to another in the joint coordinate frame, the robot arm motion still generates reaction forces/moments which in turn create oscillations on the mobile platform. The states of the upper cable tensions are stabilized in less than 0.2 seconds. Because of the action of the PID controller, vibrations of the HCDPR are well controlled in this case. 

\begin{itemize}
\item Case 4(b): The Robot Arm is Fixed and the Mobile Platform is Moving
\end{itemize}\par
In this case, the robot arm is fixed and the mobile platform is moving. When the PID controller is applied ($K_p = 5 \times 10^5, K_i = 3.5 \times 10^7, {~}{\rm{and}}{~} K_d = 1.1 \times 10^4$), the desired trajectories of the mobile platform are as follows

\begin{align}
\left\{ \begin{array}{l}
{\color{black}x_m} =  - 0.1t,\quad t \in \left[ {0,{t_{\max }}} \right]\\
{\color{black}z_m} =  - 0.05t,\quad t \in \left[ {0,{t_{\max }}} \right]\\
{{\color{black}\theta_m}} = 0\\
{\color{black}\theta_{1}} = 0\\
{\color{black}\theta_{2}} = 0
\end{array} \right.
\label{eq:J5_4-4}
\end{align}
where $t$ and ${t_{\max }}$ are the current and maximum running time.

\begin{figure}[!t]
\centering
\subfigure[]{\label{subfig:J5_Case_4b_a}\includegraphics[width=42mm]{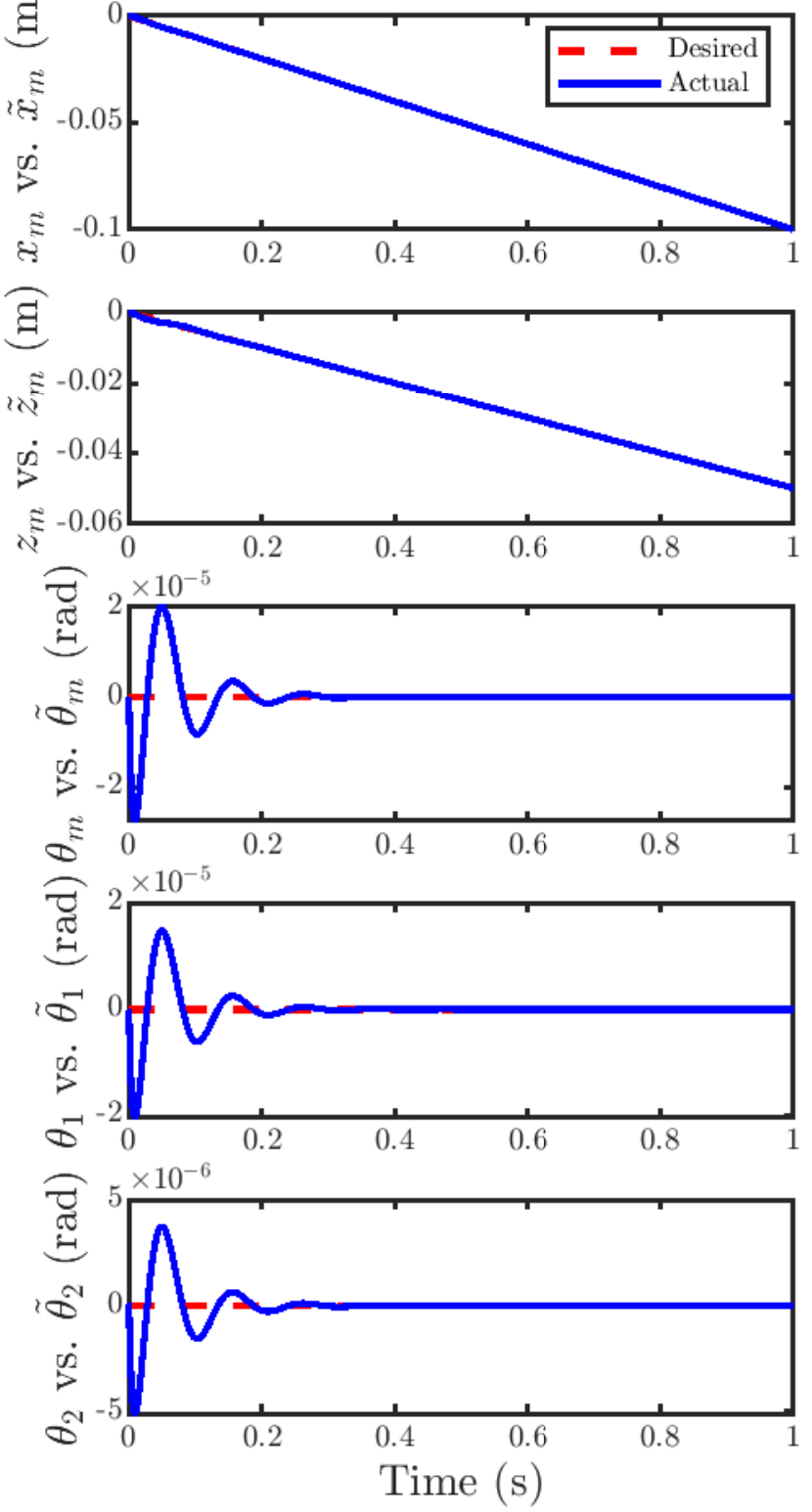}}
\subfigure[]{\label{subfig:J5_Case_4b_b}\includegraphics[width=42mm]{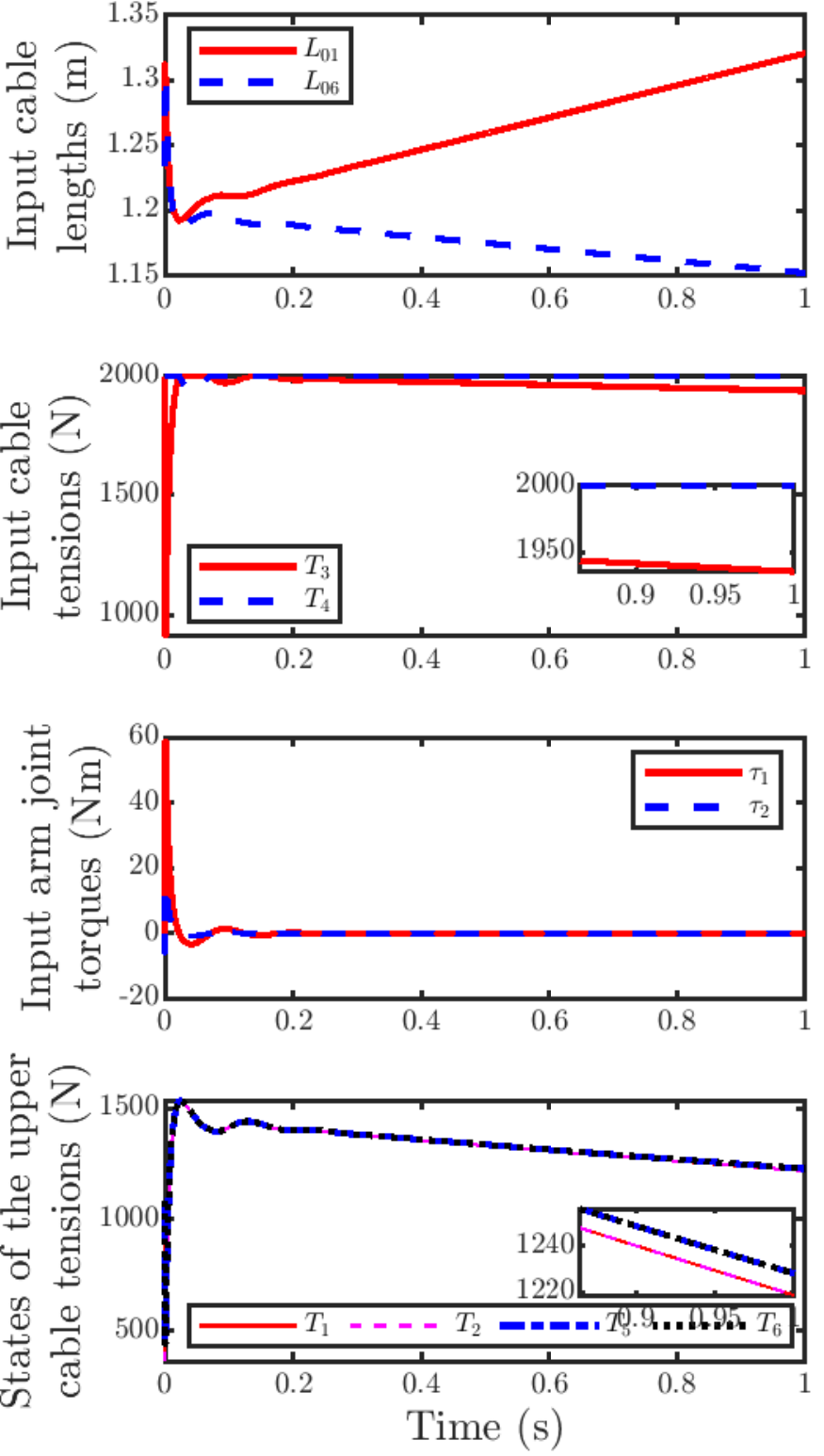}}\\
\subfigure[]{\label{subfig:J5_Case_4b_c}\includegraphics[width=84mm]{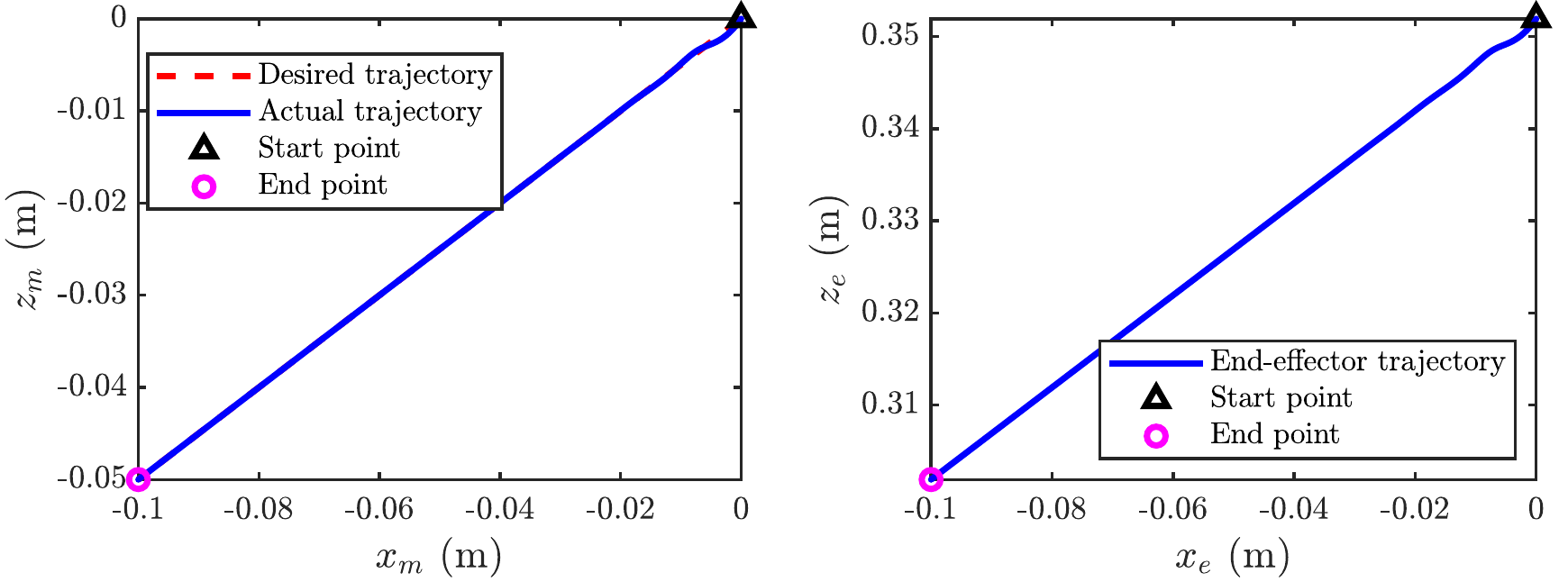}}\\
\caption{Responses of Case 4(b). (a) Errors between the desired input ${\bf{q}}$ and the actual output ${\bf{\tilde q}}$, (b) the dynamic inputs (cable lengths, cable tensions, and robot arm joint torques) applied to the HCDPR and the states of the upper cable tensions, and (c) trajectories of the center of mass of the mobile platform and the end-effector.}
\label{fig:J5_Case_4b}
\end{figure}

The results are shown in Figure~\ref{subfig:J5_Case_4b_a}, Figure~\ref{subfig:J5_Case_4b_b}, and Figure~\ref{subfig:J5_Case_4b_c}. The results show that the errors between the desired input ${\bf{q}}$ and the actual output ${\bf{\tilde q}}$ go to zero in about 0.25 seconds. The dynamic inputs (cable lengths, cable tensions, and robot arm joint torques) applied to the HCDPR are also quick to stabilize. Meanwhile, cable tensions $T_3$ and $T_4$ are always positive since the algorithm for maximizing the stiffness of HCDPR is applied. In this case, the upper cable tensions reach the set values in about 0.2 seconds. The tracking trajectory errors of the center of mass of the mobile platform shown in Figure~\ref{subfig:J5_Case_4b_c} are very small. In addition, when the proposed PID controller is implemented, the vibrations of the HCDPR are well controlled.

In summary, redundancy resolution and stiffness optimization methods for the HCDPR were introduced. PID-based controllers are also designed for position control of the HCDPR system. The performance of the HCDPR using the position PID controllers is analyzed via different scenarios: when the positions/orientations of the mobile platform and the end-effector positions of the rigid robot arm (or joint variables) are given, the trajectory tracking and vibration suppression can be well handled.

\section{Conclusions}
\label{sec:J5_Conclusions}
This paper proposed a kinematically constrained planar HCDPR which can harness the strengths and benefits of serial and cable-driven parallel robots. Based on this HCDPR, kinematics, dynamics, redundancy resolution and stiffness maximization algorithms were developed. Controllers (I and II) were also designed to address trajectory tracking and vibration suppression problems. Control performance was analyzed by using different scenarios, and the results showed that the controller II can achieve the goal better. Besides, compared to the existing research, this paper showed the reaction performance, i.e., the mobile platform was fixed and the robot arm was moving or the mobile platform was fixed and the robot arm is moving, as well as the trajectory tracking of the end-effector, and both results were satisfactory.

\section*{Acknowledgment}
The authors would like to knowledge the financial support of the Natural Sciences and Engineering Research Council of Canada (NSERC).

\section*{Appendix A: HCDPR Derivations} \label{appendix:J5_HCDPR_Drivations}
The terms in \eqref{eq:J5_20} are computed as follows:

${\bf{M}}({\bf{q}}) = \begin{bmatrix}
{{M_{11}}}&{{M_{12}}}&{{M_{13}}}&{{M_{14}}}&{{M_{15}}}\\
{{M_{21}}}&{{M_{22}}}&{{M_{23}}}&{{M_{24}}}&{{M_{25}}}\\
{{M_{31}}}&{{M_{32}}}&{{M_{33}}}&{{M_{34}}}&{{M_{35}}}\\
{{M_{41}}}&{{M_{42}}}&{{M_{43}}}&{{M_{44}}}&{{M_{45}}}\\
{{M_{51}}}&{{M_{52}}}&{{M_{53}}}&{{M_{54}}}&{{M_{55}}}
\end{bmatrix}$, ${\bf{C}}( {{\bf{q,\dot q}}} ) = \begin{bmatrix}
{{C_{11}}}&{{C_{12}}}&{{C_{13}}}&{{C_{14}}}&{{C_{15}}}\\
{{C_{21}}}&{{C_{22}}}&{{C_{23}}}&{{C_{24}}}&{{C_{25}}}\\
{{C_{31}}}&{{C_{32}}}&{{C_{33}}}&{{C_{34}}}&{{C_{35}}}\\
{{C_{41}}}&{{C_{42}}}&{{C_{43}}}&{{C_{44}}}&{{C_{45}}}\\
{{C_{51}}}&{{C_{52}}}&{{C_{53}}}&{{C_{54}}}&{{C_{55}}}
\end{bmatrix}$, ${\bf{G}}( {\bf{q}} ) = {\begin{bmatrix}
{{G_1}}\\{{G_2}}\\{{G_3}}\\{{G_4}}\\{{G_5}}
\end{bmatrix}}$, and ${{\bf{P}}_{vs}}( {\bf{q}} ) = {\begin{bmatrix}
{{P_{vs1}}}&{{P_{vs2}}}&{{P_{vs3}}}&0&0
\end{bmatrix}^T}$, in which ${M_{11}} = {m_1} + {m_2} + {m_m}$, ${M_{21}} = 0$, ${M_{31}} =  - {l_m} {m_1} \sin ( {{{\color{black}\theta_m}}} ) - {l_m} {m_2} \sin ( {{{\color{black}\theta_m}}} ) - {l_{c2}} {m_2} \sin ( {{{\color{black}\theta_m}} + {\color{black}\theta_{1}} + {\color{black}\theta_{2}}} ) - {l_1} {m_2} \sin ( {{{\color{black}\theta_m}} + {\color{black}\theta_{1}}} )
 - {l_{c1}} {m_1} \sin ( {{{\color{black}\theta_m}} + {\color{black}\theta_{1}}} )$, ${M_{41}} =  - {l_{c2}} {m_2} \sin ( {{{\color{black}\theta_m}} + {\color{black}\theta_{1}} + {\color{black}\theta_{2}}} ) - {l_1} {m_2} \sin ( {{{\color{black}\theta_m}} + {\color{black}\theta_{1}}} ) - {l_{c1}} {m_1} \sin ( {{{\color{black}\theta_m}} + {\color{black}\theta_{1}}} )$, $
{M_{51}} =  - {l_{c2}} {m_2} \sin ( {{\color{black}\theta_m} + {\color{black}\theta_{1}} + {\color{black}\theta_{2}}} )$, ${M_{12}} = 0$, ${M_{22}} = {m_1} + {m_2} + {m_m}$, $
{M_{32}} = {l_m} {m_1} \cos ( {{\color{black}\theta_m}} ) + {l_m} {m_2} \cos ( {{\color{black}\theta_m}} ) + {l_{c2}} {m_2} \cos ( {{\color{black}\theta_m} + {\color{black}\theta_{1}} + {\color{black}\theta_{2}}} ) + {l_1} {m_2} \cos ( {{\color{black}\theta_m} + {\color{black}\theta_{1}}} ) + {l_{c1}} {m_1} \cos ( {{\color{black}\theta_m} + {\color{black}\theta_{1}}} )$, $
{M_{42}} = {l_{c2}} {m_2} \cos ( {{\color{black}\theta_m} + {\color{black}\theta_{1}} + {\color{black}\theta_{2}}} ) + {l_1} {m_2} \cos ( {{\color{black}\theta_m} + {\color{black}\theta_{1}}} ) + {l_{c1}} {m_1} \cos ( {{\color{black}\theta_m} + {\color{black}\theta_{1}}} )$, $
{M_{52}} = {l_{c2}} {m_2} \cos ( {{\color{black}\theta_m} + {\color{black}\theta_{1}} + {\color{black}\theta_{2}}} )$, ${M_{13}} =  - {m_2} ({l_1} \sin ( {{\color{black}\theta_m} +{\color{black}\theta_{1}}} ) +  {l_m} \sin ( {{\color{black}\theta_m}} ) +  {l_{c2}} \sin ( {{\color{black}\theta_m} + {\color{black}\theta_{1}} +{\color{black}\theta_{2}}} )) - {m_1} ( { {l_{c1}} \sin ( {{\color{black}\theta_m} + {\color{black}\theta_{1}}} ) +  {l_m} \sin ( {{\color{black}\theta_m}} )} )$, ${M_{23}} = {m_2} ({l_1} \cos ( {{\color{black}\theta_m} + {\color{black}\theta_{1}}} ) +  {l_m} \cos ( {{\color{black}\theta_m}} ) +  {l_{c2}} \cos ( {{\color{black}\theta_m} + {\color{black}\theta_{1}} + {\color{black}\theta_{2}}} )) + {m_1} ( {l_{c1}} \cos ( {{\color{black}\theta_m} + {\color{black}\theta_{1}}} ) +  {l_m} \cos ( {{\color{black}\theta_m}} ))$, ${M_{33}} = {I_1} + {I_2} + {I_m} + {l_1}^2 {m_2} + {l_{c1}}^2 {m_1} + {l_{c2}}^2 {m_2} + {l_m}^2 {m_1} + {l_m}^2 {m_2} + 2 {l_{c2}} {l_m} {m_2} \cos ( {{\color{black}\theta_{1}} + {\color{black}\theta_{2}}} )
 + 2 {l_1} {l_{c2}} {m_2} \cos ( {{\color{black}\theta_{2}}} ) + 2 {l_1} {l_m} {m_2} \cos ( {{\color{black}\theta_{1}}} ) + 2 {l_{c1}} {l_m} {m_1} \cos ( {{\color{black}\theta_{1}}} )$, ${M_{43}} = {m_2} {l_1}^2 + 2 {m_2} \cos ( {{\color{black}\theta_{2}}} ) {l_1} {l_{c2}} + {l_m} {m_2} \cos ( {{\color{black}\theta_{1}}} ) {l_1} + {m_1} {l_{c1}}^2 + {l_m} {m_1} \cos ( {{\color{black}\theta_{1}}} ) {l_{c1}} + {m_2} {l_{c2}}^2 + {l_m} {m_2} \cos ( {{\color{black}\theta_{1}} + {\color{black}\theta_{2}}} ) {l_{c2}} + {I_1} + {I_2}$, ${M_{53}} = {I_2} + {l_{c2}}^2 {m_2} + {l_{c2}} {l_m} {m_2} \cos ( {{\color{black}\theta_{1}} + {\color{black}\theta_{2}}} ) + {l_1} {l_{c2}} {m_2} \cos ( {{\color{black}\theta_{2}}} )$, ${M_{14}} =  - {m_2} ( { {l_1} \sin ( {{\color{black}\theta_m} + {\color{black}\theta_{1}}} ) +  {l_{c2}} \sin ( {{\color{black}\theta_m} + {\color{black}\theta_{1}} + {\color{black}\theta_{2}}} )} ) - {l_{c1}} {m_1} \sin ( {{\color{black}\theta_m} + {\color{black}\theta_{1}}} )$, ${M_{24}} = {m_2} ({l_1} \cos ( {{\color{black}\theta_m} + {\color{black}\theta_{1}}} ) +  {l_{c2}} \cos ( {{\color{black}\theta_m} + {\color{black}\theta_{1}} + {\color{black}\theta_{2}}} )) + {l_{c1}} {m_1} \cos ( {{\color{black}\theta_m} + {\color{black}\theta_{1}}} )$, ${M_{34}} = {m_2} {l_1}^2 + 2 {m_2} \cos ( {{\color{black}\theta_{2}}} ) {l_1} {l_{c2}} + {l_m} {m_2} \cos ( {{\color{black}\theta_{1}}} ) {l_1} + {m_1} {l_{c1}}^2 + {l_m} {m_1} \cos ( {{\color{black}\theta_{1}}} ) {l_{c1}} + {m_2} {l_{c2}}^2 + {l_m} {m_2} \cos ( {{\color{black}\theta_{1}} + {\color{black}\theta_{2}}} ) {l_{c2}} + {I_1} + {I_2}$, ${M_{44}} = {m_2} {l_1}^2 + 2 {m_2} \cos ( {{\color{black}\theta_{2}}} ) {l_1} {l_{c2}} + {m_1} {l_{c1}}^2 + {m_2} {l_{c2}}^2 + {I_1} + {I_2}$, ${M_{54}} = {m_2} {l_{c2}}^2 + {l_1} {m_2} \cos ( {{\color{black}\theta_{2}}} ) {l_{c2}} + {I_2}$, ${M_{15}} =  - {l_{c2}} {m_2} \sin ( {{\color{black}\theta_m} + {\color{black}\theta_{1}} + {\color{black}\theta_{2}}} )$, ${M_{25}} = {l_{c2}} {m_2} \cos ( {{\color{black}\theta_m} + {\color{black}\theta_{1}} + {\color{black}\theta_{2}}} )$, ${M_{35}} = {I_2} + {l_{c2}}^2 {m_2} + {l_{c2}} {l_m} {m_2} \cos ( {{\color{black}\theta_{1}} + {\color{black}\theta_{2}}} ) + {l_1} {l_{c2}} {m_2} \cos ( {{\color{black}\theta_{2}}} )$, ${M_{45}} = {m_2} {l_{c2}}^2 + {l_1} {m_2} \cos ( {{\color{black}\theta_{2}}} ) {l_{c2}} + {I_2}$, ${M_{55}} = {l_{c2}}^2 {m_2} + {I_2}$, ${C_{11}} = 0$, ${C_{12}} = 0$, ${C_{13}} =  - {{\dot \theta}_m} ({m_2} ({l_1} \cos ( {\color{black}\theta_m} + {\color{black}\theta_{1}}) +  {l_m} \cos ({{\color{black}\theta_m}} ) +  {l_{c2}} \cos ({\color{black}\theta_m} + {\color{black}\theta_{1}} + {\color{black}\theta_{2}})) + {m_1} ({l_{c1}} \cos ({\color{black}\theta_m} + {\color{black}\theta_{1}}) +  {l_m} \cos ( {{\color{black}\theta_m}} )))$, ${C_{14}} =  - (2 {{\dot \theta}_m} + {{\dot \theta}_{1}}) ( {l_{c2}} {m_2} \cos ( {{\color{black}\theta_m} + {\color{black}\theta_{1}} + {\color{black}\theta_{2}}} ) + {l_1} {m_2} \cos ( {{\color{black}\theta_m} + {\color{black}\theta_{1}}} ) + {l_{c1}} {m_1} \cos ( {{\color{black}\theta_m} + {\color{black}\theta_{1}}} ))$, $
{C_{15}} =  - {l_{c2}} {m_2} \cos ({\color{black}\theta_m} + {\color{black}\theta_{1}} + {\color{black}\theta_{2}}) (2 {{\dot \theta}_m} + 2 {{\dot \theta}_{1}} + {{\dot \theta}_{2}})$, ${C_{21}} = 0$, ${C_{22}} = 0$, $
{C_{23}} =  - {{\dot \theta}_m} ({m_2} ({l_1} \sin ({\color{black}\theta_m} + {\color{black}\theta_{1}}) +  {l_m} \sin ({{\color{black}\theta_m}} ) +  {l_{c2}} \sin( {\color{black}\theta_m} + {\color{black}\theta_{1}} + {\color{black}\theta_{2}})) + {m_1} ({l_{c1}} \sin ({\color{black}\theta_m} + {\color{black}\theta_{1}}) +  {l_m} \sin ( {{\color{black}\theta_m}} )))$, ${C_{24}} =  - ({2 {{\dot \theta}_m} + {{\dot \theta}_{1}}} ) ({l_{c2}} {m_2} \sin ( {{\color{black}\theta_m} + {\color{black}\theta_{1}} + {\color{black}\theta_{2}}} ) + {l_1} {m_2} \sin ( {{\color{black}\theta_m} + {\color{black}\theta_{1}}} ) + {l_{c1}} {m_1} \sin ({\color{black}\theta_m} + {\color{black}\theta_{1}}))$, $
{C_{25}} =  - {l_{c2}} {m_2} \sin ({\color{black}\theta_m} + {\color{black}\theta_{1}} + {\color{black}\theta_{2}}) (2 {{\dot \theta}_m} + 2 {{\dot \theta}_{1}} + {{\dot \theta}_{2}})$, ${C_{31}} = 0$, ${C_{32}} = 0$, ${C_{33}} = 0$, $
{C_{34}} =  - {l_m} (2 {{\dot \theta}_m} + {{\dot \theta}_{1}}) ({l_1} {m_2} \sin ( {{\color{black}\theta_{1}}} ) + {l_{c1}} {m_1} \sin ({{\color{black}\theta_{1}}}) + {l_{c2}} {m_2} \sin ( {{\color{black}\theta_{1}} + {\color{black}\theta_{2}}} ))$, $
{C_{35}} =  - {l_{c2}} {m_2} ( {{l_m} \sin ( {{\color{black}\theta_{1}} + {\color{black}\theta_{2}}} ) + {l_1} \sin ( {{\color{black}\theta_{2}}} )} ) ( {2 {{\dot \theta}_m} + 2 {{\dot \theta}_{1}} + {{\dot \theta}_{2}}} )$, 
${C_{41}} = 0$, ${C_{42}} = 0$, ${C_{43}} = {l_m} {{\dot \theta}_m} ( {{l_1} {m_2} \sin ( {{\color{black}\theta_{1}}} ) + {l_{c1}} {m_1} \sin ( {{\color{black}\theta_{1}}} ) + {l_{c2}} {m_2} \sin ( {{\color{black}\theta_{1}} + {\color{black}\theta_{2}}} )} )$, $
{C_{44}} = 0$, ${C_{45}} =  - {l_1} {l_{c2}} {m_2} \sin ( {{\color{black}\theta_{2}}} ) ( {2 {{\dot \theta}_m} + 2 {{\dot \theta}_{1}} + {{\dot \theta}_{2}}} )$, ${C_{51}} = 0$, ${C_{52}} = 0$, $
{C_{53}} = {l_{c2}} {m_2} {{\dot \theta}_m} ( {{l_m} \sin ( {{\color{black}\theta_{1}} + {\color{black}\theta_{2}}} ) + {l_1} \sin ( {{\color{black}\theta_{2}}} )} )$, ${C_{54}} = {l_1} {l_{c2}} {m_2} \sin ( {{\color{black}\theta_{2}}} ) ( {2 {{\dot \theta}_m} + {{\dot \theta}_{1}}} )$, ${C_{55}} = 0,$
${G_1} = 0$, ${G_2} =  ( {{m_1} + {m_2} + {m_m}} )g$, $
{G_3} =  {m_1}g{l_m}  \cos ( {{\color{black}\theta_m}} ) + {m_2}g{l_m}  \cos ( {{\color{black}\theta_m}} ) + {m_2}g{l_{c2}}  \cos ( {{\color{black}\theta_m} + {\color{black}\theta_{1}} + {\color{black}\theta_{2}}} ) + {m_2}g{l_1}  \cos ( {{\color{black}\theta_m} + {\color{black}\theta_{1}}} )
 + {m_1}g{l_{c1}}  \cos ( {{\color{black}\theta_m} + {\color{black}\theta_{1}}} )$, $
{G_4} =   {m_2}g{l_1} \cos ( {{\color{black}\theta_m} + {\color{black}\theta_{1}}} ) +   {m_1} g{l_{c1}}\cos ( {{\color{black}\theta_m} + {\color{black}\theta_{1}}} ) +   {m_2}g{l_{c2}} \cos ( {{\color{black}\theta_m} + {\color{black}\theta_{1}} + {\color{black}\theta_{2}}} )$, $
{G_5} = {m_2}g {l_{c2}}  \cos ( {{\color{black}\theta_m} + {\color{black}\theta_{1}} + {\color{black}\theta_{2}}} )$, 
${P_{vs1}} = {k_x}( {{\color{black}x_m} - {\color{black}x_{m0}}} )$, ${P_{vs2}} = {k_z}( {{\color{black}z_m} - {\color{black}z_{m0}}} )$, ${P_{vs3}} = {k_\theta }( {{\color{black}\theta_m} - {\color{black}\theta_{m0}}} )$, ${P_{vs4}} = 0$, ${P_{vs5}} = 0.$

\section*{Appendix B: Equivalent Cable-Driven Model}\label{appendix:J5_EquivalentModel}
\begin{figure}[h]\centering
	\includegraphics[width=90mm]{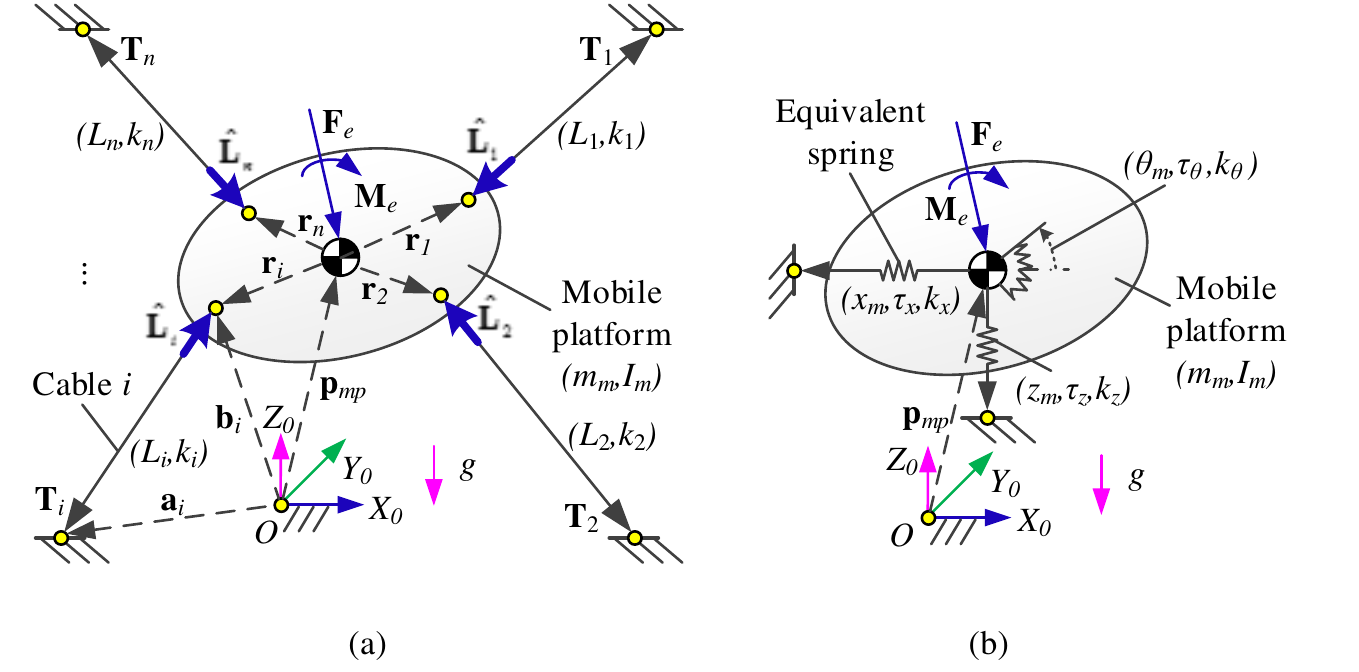}
	\caption{An equivalent three-spring driven model for a 2D flexible CDPR. (a) A 2D flexible CDPR; (b) an equivalent three-spring driven model.}\label{fig:J5_EquivalentModel}
\end{figure}

\begin{theorem}
Assume an external force and moment ${\left[ {{{\mathbf{F}}_e},{{\mathbf{M}}_e}} \right]^T} \in {\mathbb{R}^3}$ are applied to the mobile platform in a 2D CDPR (shown in \autoref{fig:J5_EquivalentModel}), then the equation ${{\mathbf{\tau }}_m} =  - {\mathbf{AT}}$ will be satisfied. In this equation, ${{\mathbf{\tau }}_m}: = {\left[ {{\tau _x},{\tau _z},{\tau _\theta }} \right]^T} \in {\mathbb{R}^3},$ ${\mathbf{A}} \in {\mathbb{R}^{3 \times n}},$ and ${\mathbf{T}}: = {\left[ {{T_1},{T_2}, \cdots ,{T_n}} \right]^T} \in {\mathbb{R}^n}$ represent the equivalent joint forces/torques applied to the mobile platform, the structure matrix $\bf{A}$, and the cable tensions, respectively. In \autoref{fig:J5_EquivalentModel}(b), suppose ${\tau _x},{\tau _z},$ and ${\tau _\theta }$ always parallel axes $O{X_0}$, $O{Z_0}$, and $O{Y_0}$, respectively.
\end{theorem}

\begin{proof}
Suppose an external force and moment ${\left[ {{{\mathbf{F}}_e},{{\mathbf{M}}_e}} \right]^T} \in {\mathbb{R}^3}$ are applied to the mobile platform (as shown in \autoref{fig:J5_EquivalentModel}(a) and \autoref{fig:J5_EquivalentModel}(b)) and generate the same position and orientation accelerations ${\left[ {{{\ddot x}_m},{{\ddot z}_m},{{\ddot \theta }_m}} \right]^T}$. Using the Newton-Euler formula, the following equations can be derived.

For the model shown in \autoref{fig:J5_EquivalentModel}(a), we have
\begin{align}
\sum\limits_{i = 1}^n {{{\bf{T}}_i}}  + {{\bf{F}}_e} + \begin{bmatrix}
0\\
{{m_m}g}
\end{bmatrix} &= \begin{bmatrix}
{{m_m}{{\ddot x}_m}}\\
{{m_m}{{\ddot z}_m}}
\end{bmatrix}\nonumber\\

 - \sum\limits_{i = 1}^n {\left( {{{{\bf{\hat L}}}_i}{T_i}} \right)}  &= \begin{bmatrix}
{{m_m}{{\ddot x}_m}}\\
{{m_m}{{\ddot z}_m}}
\end{bmatrix} - \begin{bmatrix}
0\\
{{m_m}g}
\end{bmatrix} - {{\bf{F}}_e}
\label{eq:J5_appendix_A_1}
\end{align}
where ${{\bf{\hat L}}_i}$ denotes the unit cable vector. Furthermore, 
\begin{align}
\sum\limits_{i = 1}^n {\left( {{{\bf{r}}_i} \times {{\bf{T}}_i}} \right)}  + {{\bf{M}}_e} &= {I_m}{{\ddot \theta }_m}\nonumber\\

 - \sum\limits_{i = 1}^n {\left( {\left( {{{\bf{r}}_i} \times {{{\bf{\hat L}}}_i}} \right){T_i}} \right)} &= {I_m}{{\ddot \theta }_m} - {{\bf{M}}_e}
\label{eq:J5_appendix_A_2}
\end{align}

Combining \eqref{eq:J5_appendix_A_1} and \eqref{eq:J5_appendix_A_2}, we get
\begin{align}
 - \sum\limits_{i = 1}^n {\left\{ {\begin{bmatrix}
{{{{\bf{\hat L}}}_i}}\\
{{{\bf{r}}_i} \times {{{\bf{\hat L}}}_i}}
\end{bmatrix}{T_i}} \right\}}  = \begin{bmatrix}
{{m_m}{{\ddot x}_m}}\\
{{m_m}{{\ddot z}_m}}\\
{{I_m}{{\ddot \theta }_m}}
\end{bmatrix} - \begin{bmatrix}
0\\
{{m_m}g}\\
0
\end{bmatrix} - \begin{bmatrix}
{{{\bf{F}}_e}}\\
{{{\bf{M}}_e}}
\end{bmatrix}
\label{eq:J5_appendix_A_3}
\end{align}

For the model shown in \autoref{fig:J5_EquivalentModel}(b), we also have
\begin{align}
\begin{bmatrix}
{{\tau _x}}\\
{{\tau _z}}
\end{bmatrix} &= \begin{bmatrix}
{{m_m}{{\ddot x}_m}}\\
{{m_m}{{\ddot z}_m}}
\end{bmatrix} - \begin{bmatrix}
0\\
{{m_m}g}
\end{bmatrix} - {{\bf{F}}_e}
\label{eq:J5_appendix_A_4}
\end{align}
and
\begin{align}
{\tau _\theta } &= {I_m}{{\ddot \theta }_m} - {{\bf{M}}_e}
\label{eq:J5_appendix_A_5}
\end{align}

Combining \eqref{eq:J5_appendix_A_4} and \eqref{eq:J5_appendix_A_5}, we get
\begin{align}
\begin{bmatrix}
{{\tau _x}}\\
{{\tau _z}}\\
{{\tau _\theta }}
\end{bmatrix} = \begin{bmatrix}
{{m_m}{{\ddot x}_m}}\\
{{m_m}{{\ddot z}_m}}\\
{{I_m}{{\ddot \theta }_m}}
\end{bmatrix} - \begin{bmatrix}
0\\
{{m_m}g}\\
0
\end{bmatrix} - \begin{bmatrix}
{{{\bf{F}}_e}}\\
{{{\bf{M}}_e}}
\end{bmatrix}
\label{eq:J5_appendix_A_6}
\end{align}

Clearly, the right sides of \eqref{eq:J5_appendix_A_3} and \eqref{eq:J5_appendix_A_6} are equal, so
\begin{align}
\begin{bmatrix}
{{\tau _x}}\\
{{\tau _z}}\\
{{\tau _\theta }}
\end{bmatrix} =  - \sum\limits_{i = 1}^n {\left\{ {\begin{bmatrix}
{{{{\bf{\hat L}}}_i}}\\
{{{\bf{r}}_i} \times {{{\bf{\hat L}}}_i}}
\end{bmatrix}{T_i}} \right\}}
\label{eq:J5_appendix_A_7}
\end{align}

Eq. \eqref{eq:J5_appendix_A_7} is expanded as
\begin{align}
{{\bf{\tau }}_m}= - \underbrace {\begin{bmatrix}
{{{{\bf{\hat L}}}_1}}&{{{{\bf{\hat L}}}_2}}& \cdots &{{{{\bf{\hat L}}}_i}}& \cdots &{{{{\bf{\hat L}}}_n}}\\
{{{\bf{r}}_1} \times {{{\bf{\hat L}}}_1}}&{{{\bf{r}}_2} \times {{{\bf{\hat L}}}_2}}& \cdots &{{{\bf{r}}_i} \times {{{\bf{\hat L}}}_i}}& \cdots &{{{\bf{r}}_n} \times {{{\bf{\hat L}}}_n}}
\end{bmatrix}}_A{\bf{T}}
\label{eq:J5_appendix_A_8}
\end{align}
where ${{\mathbf{\hat L}}_i} = {\left[ {{{{\mathbf{\hat L}}}_{ix}},{{{\mathbf{\hat L}}}_{iz}}} \right]^T} \in {\mathbb{R}^2}$, ${{\mathbf{r}}_i} = {\left[ {{{\mathbf{r}}_{ix}},{{\mathbf{r}}_{iz}}} \right]^T} \in {\mathbb{R}^2}$, and ${\bf{T}}={\begin{bmatrix} {{T_1}}&{{T_2}}& \cdots &{{T_i}}& \cdots &{{T_n}} \end{bmatrix}^T} \in {\mathbb{R}^n}$. Hence, we get
\begin{align}
{{\bf{\tau }}_m} =  - {\bf{AT}}
\label{eq:J5_appendix_A_9}
\end{align}
where $\bf{A}$ represents a structure matrix, determined by the position and orientation of the mobile platform.

Furthermore, ${{\bf{\tau }}_m}$ is satisfied with ${{\bf{\tau }}_m} = {\left[ {{\tau _x},{\tau _z},{\tau _\theta }} \right]^T} = {\left[ {{k_x}\left( {{x_m} - {x_{m0}}} \right),{k_z}\left( {{z_m} - {z_{m0}}} \right),{k_\theta }\left( {{\theta _m} - {\theta _{m0}}} \right)} \right]^T}$, where ${k_x},{k_z},{k_\theta } \in \mathbb{R}$ denote equivalent spring constants (parallel the X-axis, Z-axis, and rotation about Y-axis, respectively), $\left( {{x_m},{z_m},{\theta _m}} \right)$ and $\left( {{x_{m0}},{z_{m0}},{\theta _{m0}}} \right)$ represent the current and initial positions and orientation of the mobile platform, respectively. For the six-cable HCDPR shown in \autoref{fig:J5_PlanarModel}. Then, we also get
\begin{align}
{\bf{AT}} =  - {{\bf{\tau }}_m}
 =  - {\left[ {{k_x}\left( {{x_m} - {x_{m0}}} \right),{k_z}\left( {{z_m} - {z_{m0}}} \right),{k_\theta }\left( {{\theta _m} - {\theta _{m0}}} \right)} \right]^T}
\label{eq:J5_appendix_A_10}
\end{align}

Besides, suppose ${T_i} = \left\{ {\begin{array}{*{20}{l}}
{{k_i}({L_i} - {L_{0i}})}\\
{{T_i}}
\end{array}} \right.\quad \begin{array}{*{20}{c}}
{{\rm{input}}\;{\rm{is}}\;{\rm{the}}\;i{\rm{th}}\;{\rm{cable}}\;{\rm{length}}\;}\\
{{\rm{ input}}\;{\rm{is}}\;{\rm{the}}\;i{\rm{th}}\;{\rm{cable}}\;{\rm{tension}}}
\end{array}$. It is clear that \eqref{eq:J5_appendix_A_9} is available to the cable position (cable length) control, force (cable tension) control, and hybrid cable position/force control.
\end{proof}

\section*{Appendix C: Derivations of the Maximizing Stiffness of the HCDPR}\label{appendix:J5_SomeTermsStiffness}
\begin{align}
{{\bf{D}}_A} = {\begin{bmatrix}
{{D_{A1}}}&{{D_{A2}}}&{{D_{A3}}}&{{D_{A4}}}&{{D_{A5}}}&{{D_{A6}}}
\end{bmatrix}^T}
\label{eq:J5_appendix_B_1}
\end{align}
where
\[\begin{array}{l}
{D_{A1}} ={N}_{A13}\\
{\quad\quad\;\;} - \frac{{{N_{A12}}\left( {\left( {{N}_{A11} - {N}_{A21}} \right)\left( {{N}_{A53} - {N}_{A63}} \right) - \left( {{N}_{A13} - {N}_{A23}} \right)\left( {{N}_{A51} - {N}_{A61}} \right)} \right)}}{{\left( {{N}_{A11} - {N}_{A21}} \right)\left( {{N}_{A52} - {N}_{A62}} \right) - \left( {{N}_{A12} - {N}_{A22}} \right)\left( {{N}_{A51} - {N}_{A61}} \right)}}\\
{\quad\quad\;\;}+ \frac{{{N}_{A11}\left( {\left( {{N}_{A12} - {N}_{A22}} \right)\left( {{N}_{A53} - {N}_{A63}} \right) - \left( {{N}_{A13} - {N}_{A23}} \right)\left( {{N}_{A52} - {N}_{A62}} \right)} \right)}}{{\left( {{N}_{A11} - {N}_{A21}} \right)\left( {{N}_{A52} - {N}_{A62}} \right) - \left( {{N}_{A12} - {N}_{A22}} \right)\left( {{N}_{A51} - {N}_{A61}} \right)}}\\
{D_{A2}} ={N}_{A23}\\
{\quad\quad\;\;} - \frac{{{N}_{A22}\left( {\left( {{N}_{A11} - {N}_{A21}} \right)\left( {{N}_{A53} - {N}_{A63}} \right) - \left( {{N}_{A13} - {N}_{A23}} \right)\left( {{N}_{A51} - {N}_{A61}} \right)} \right)}}{{\left( {{N}_{A11} - {N}_{A21}} \right)\left( {{N}_{A52} - {N}_{A62}} \right) - \left( {{N}_{A12} - {N}_{A22}} \right)\left( {{N}_{A51} - {N}_{A61}} \right)}}\\
{\quad\quad\;\;}+ \frac{{{N}_{A21}\left( {\left( {{N}_{A12} - {N}_{A22}} \right)\left( {{N}_{A53} - {N}_{A63}} \right) - \left( {{N}_{A13} - {N}_{A23}} \right)\left( {{N}_{A52} - {N}_{A62}} \right)} \right)}}{{\left( {{N}_{A11} - {N}_{A21}} \right)\left( {{N}_{A52} - {N}_{A62}} \right) - \left( {{N}_{A12} - {N}_{A22}} \right)\left( {{N}_{A51} - {N}_{A61}} \right)}}\\
{D_{A3}} ={N}_{A33}\\
{\quad\quad\;\;} - \frac{{{N}_{A32}\left( {\left( {{N}_{A11} - {N}_{A21}} \right)\left( {{N}_{A53} - {N}_{A63}} \right) - \left( {{N}_{A13} - {N}_{A23}} \right)\left( {{N}_{A51} - {N}_{A61}} \right)} \right)}}{{\left( {{N}_{A11} - {N}_{A21}} \right)\left( {{N}_{A52} - {N}_{A62}} \right) - \left( {{N}_{A12} - {N}_{A22}} \right)\left( {{N}_{A51} - {N}_{A61}} \right)}}\\
{\quad\quad\;\;}+ \frac{{{N}_{A31}\left( {\left( {{N}_{A12} - {N}_{A22}} \right)\left( {{N}_{A53} - {N}_{A63}} \right) - \left( {{N}_{A13} - {N}_{A23}} \right)\left( {{N}_{A52} - {N}_{A62}} \right)} \right)}}{{\left( {{N}_{A11} - {N}_{A21}} \right)\left( {{N}_{A52} - {N}_{A62}} \right) - \left( {{N}_{A12} - {N}_{A22}} \right)\left( {{N}_{A51} - {N}_{A61}} \right)}}\\
{D_{A4}} ={N}_{A43}\\
{\quad\quad\;\;} - \frac{{{N}_{A42}\left( {\left( {{N}_{A11} - {N}_{A21}} \right)\left( {{N}_{A53} - {N}_{A63}} \right) - \left( {{N}_{A13} - {N}_{A23}} \right)\left( {{N}_{A51} - {N}_{A61}} \right)} \right)}}{{\left( {{N}_{A11} - {N}_{A21}} \right)\left( {{N}_{A52} - {N}_{A62}} \right) - \left( {{N}_{A12} - {N}_{A22}} \right)\left( {{N}_{A51} - {N}_{A61}} \right)}}\\
{\quad\quad\;\;}+ \frac{{{N}_{A41}\left( {\left( {{N}_{A12} - {N}_{A22}} \right)\left( {{N}_{A53} - {N}_{A63}} \right) - \left( {{N}_{A13} - {N}_{A23}} \right)\left( {{N}_{A52} - {N}_{A62}} \right)} \right)}}{{\left( {{N}_{A11} - {N}_{A21}} \right)\left( {{N}_{A52} - {N}_{A62}} \right) - \left( {{N}_{A12} - {N}_{A22}} \right)\left( {{N}_{A51} - {N}_{A61}} \right)}}\\
{D_{A5}} ={N}_{A53}\\
{\quad\quad\;\;} - \frac{{{N}_{A52}\left( {\left( {{N}_{A11} - {N}_{A21}} \right)\left( {{N}_{A53} - {N}_{A63}} \right) - \left( {{N}_{A13} - {N}_{A23}} \right)\left( {{N}_{A51} - {N}_{A61}} \right)} \right)}}{{\left( {{N}_{A11} - {N}_{A21}} \right)\left( {{N}_{A52} - {N}_{A62}} \right) - \left( {{N}_{A12} - {N}_{A22}} \right)\left( {{N}_{A51} - {N}_{A61}} \right)}}\\
{\quad\quad\;\;}+ \frac{{{N}_{A51}\left( {\left( {{N}_{A12} - {N}_{A22}} \right)\left( {{N}_{A53} - {N}_{A63}} \right) - \left( {{N}_{A13} - {N}_{A23}} \right)\left( {{N}_{A52} - {N}_{A62}} \right)} \right)}}{{\left( {{N}_{A11} - {N}_{A21}} \right)\left( {{N}_{A52} - {N}_{A62}} \right) - \left( {{N}_{A12} - {N}_{A22}} \right)\left( {{N}_{A51} - {N}_{A61}} \right)}}\\
{D_{A6}} ={N}_{A63}\\
{\quad\quad\;\;} - \frac{{{N}_{A62}\left( {\left( {{N}_{A11} - {N}_{A21}} \right)\left( {{N}_{A53} - {N}_{A63}} \right) - \left( {{N}_{A13} - {N}_{A23}} \right)\left( {{N}_{A51} - {N}_{A61}} \right)} \right)}}{{\left( {{N}_{A11} - {N}_{A21}} \right)\left( {{N}_{A52} - {N}_{A62}} \right) - \left( {{N}_{A12} - {N}_{A22}} \right)\left( {{N}_{A51} - {N}_{A61}} \right)}}\\
{\quad\quad\;\;}+ \frac{{{N}_{A61}\left( {\left( {{N}_{A12} - {N}_{A22}} \right)\left( {{N}_{A53} - {N}_{A63}} \right) - \left( {{N}_{A13} - {N}_{A23}} \right)\left( {{N}_{A52} - {N}_{A62}} \right)} \right)}}{{\left( {{N}_{A11} - {N}_{A21}} \right)\left( {{N}_{A52} - {N}_{A62}} \right) - \left( {{N}_{A12} - {N}_{A22}} \right)\left( {{N}_{A51} - {N}_{A61}} \right)}}
\end{array}\]
\begin{align}
{{\bf{E}}_A} = {\begin{bmatrix}
{{E_{A1}}}&{{E_{A2}}}&{{E_{A3}}}&{{E_{A4}}}&{{E_{A5}}}&{{E_{A6}}}
\end{bmatrix}^T}
\label{eq:J5_appendix_B_2}
\end{align}
where
\[\begin{array}{l}
{E_{A1}} = {T_A}_1 + \frac{{N_A}{{_1}_2}({{N_A}{{_1}_1} - {N_A}{{_2}_1}})( {{T_A}_6 - {T_A}_5 + {k_5}( {{L_5} - {L_6}})})}{{( {{N_A}{{_1}_1} - {N_A}{{_2}_1}})( {{N_A}{{_5}_2} - {N_A}{{_6}_2}}) - ( {{N_A}{{_1}_2} - {N_A}{{_2}_2}})( {{N_A}{{_5}_1} - {N_A}{{_6}_1}})}}\\
{\quad\quad\;\;}- \frac{{N_A}{{_1}_2}({{N_A}{{_5}_1} - {N_A}{{_6}_1}})( {{T_A}_2 - {T_A}_1 + {k_1}( {{L_1} - {L_2}}))}}{{( {{N_A}{{_1}_1} - {N_A}{{_2}_1}})( {{N_A}{{_5}_2} - {N_A}{{_6}_2}}) - ( {{N_A}{{_1}_2} - {N_A}{{_2}_2}})( {{N_A}{{_5}_1} - {N_A}{{_6}_1}})}}\\
{\quad\quad\;\;}- \frac{{N_A}{{_1}_1}{( {{N_A}{{_1}_2} - {N_A}{{_2}_2}})( {{T_A}_6 - {T_A}_5 + {k_5}( {{L_5} - {L_6}})})}}
{{( {{N_A}{{_1}_1} - {N_A}{{_2}_1}})( {{N_A}{{_5}_2} - {N_A}{{_6}_2}}) - ( {{N_A}{{_1}_2} - {N_A}{{_2}_2}})( {{N_A}{{_5}_1} - {N_A}{{_6}_1}})}}\\
{\quad\quad\;\;}+ \frac{({N_A}{{_1}_1}({{N_A}{{_5}_2} - {N_A}{{_6}_2}})( {{T_A}_2 - {T_A}_1 + {k_1}( {{L_1} - {L_2}})}))}
{{( {{N_A}{{_1}_1} - {N_A}{{_2}_1}})( {{N_A}{{_5}_2} - {N_A}{{_6}_2}}) - ( {{N_A}{{_1}_2} - {N_A}{{_2}_2}})( {{N_A}{{_5}_1} - {N_A}{{_6}_1}})}}\\
{E_{A2}} = {T_A}_2 + \frac{{N_A}{{_2}_2}({( {{N_A}{{_1}_1} - {N_A}{{_2}_1}})( {{T_A}_6 - {T_A}_5 + {k_5}( {{L_5} - {L_6}})}))}}
{{( {{N_A}{{_1}_1} - {N_A}{{_2}_1}})( {{N_A}{{_5}_2} - {N_A}{{_6}_2}}) - ( {{N_A}{{_1}_2} - {N_A}{{_2}_2}})( {{N_A}{{_5}_1} - {N_A}{{_6}_1}})}}\\
{\quad\quad\;\;}- \frac{{N_A}{{_2}_2}({( {{N_A}{{_5}_1} - {N_A}{{_6}_1}})( {{T_A}_2 - {T_A}_1 + {k_1}( {{L_1} - {L_2}})}))}}
{{( {{N_A}{{_1}_1} - {N_A}{{_2}_1}})( {{N_A}{{_5}_2} - {N_A}{{_6}_2}}) - ( {{N_A}{{_1}_2} - {N_A}{{_2}_2}})( {{N_A}{{_5}_1} - {N_A}{{_6}_1}})}}\\
{\quad\quad\;\;}- \frac{{N_A}{{_2}_1}({( {{N_A}{{_1}_2} - {N_A}{{_2}_2}})( {{T_A}_6 - {T_A}_5 + {k_5}( {{L_5} - {L_6}})}))}}{{( {{N_A}{{_1}_1} - {N_A}{{_2}_1}})( {{N_A}{{_5}_2} - {N_A}{{_6}_2}}) - ( {{N_A}{{_1}_2} - {N_A}{{_2}_2}})( {{N_A}{{_5}_1} - {N_A}{{_6}_1}})}}\\
{\quad\quad\;\;}+ \frac{{N_A}{{_2}_1}({( {{N_A}{{_5}_2} - {N_A}{{_6}_2}})( {{T_A}_2 - {T_A}_1 + {k_1}( {{L_1} - {L_2}})}))}}{{( {{N_A}{{_1}_1} - {N_A}{{_2}_1}})( {{N_A}{{_5}_2} - {N_A}{{_6}_2}}) - ( {{N_A}{{_1}_2} - {N_A}{{_2}_2}})( {{N_A}{{_5}_1} - {N_A}{{_6}_1}})}}\\
\end{array}\]
\[\begin{array}{l}
{E_{A3}} = {T_A}_3 + \frac{{N_A}{{_3}_2}( {( {{N_A}{{_1}_1} - {N_A}{{_2}_1}})( {{T_A}_6 - {T_A}_5 + {k_5}( {{L_5} - {L_6}})}))}}
{{( {{N_A}{{_1}_1} - {N_A}{{_2}_1}})( {{N_A}{{_5}_2} - {N_A}{{_6}_2}}) - ( {{N_A}{{_1}_2} - {N_A}{{_2}_2}})( {{N_A}{{_5}_1} - {N_A}{{_6}_1}})}}\\
{\quad\quad\;\;}- \frac{{N_A}{{_3}_2}({( {{N_A}{{_5}_1} - {N_A}{{_6}_1}})( {{T_A}_2 - {T_A}_1 + {k_1}( {{L_1} - {L_2}})}))}}
{{( {{N_A}{{_1}_1} - {N_A}{{_2}_1}})( {{N_A}{{_5}_2} - {N_A}{{_6}_2}}) - ( {{N_A}{{_1}_2} - {N_A}{{_2}_2}})( {{N_A}{{_5}_1} - {N_A}{{_6}_1}})}}\\
{\quad\quad\;\;}- \frac{{N_A}{{_3}_1}({( {{N_A}{{_1}_2} - {N_A}{{_2}_2}})({{T_A}_6 - {T_A}_5 + {k_5}( {{L_5} - {L_6}})}))}}
{{( {{N_A}{{_1}_1} - {N_A}{{_2}_1}})( {{N_A}{{_5}_2} - {N_A}{{_6}_2}}) - ( {{N_A}{{_1}_2} - {N_A}{{_2}_2}})( {{N_A}{{_5}_1} - {N_A}{{_6}_1}})}}\\
{\quad\quad\;\;}+ \frac{{N_A}{{_3}_1}({( {{N_A}{{_5}_2} - {N_A}{{_6}_2}})( {{T_A}_2 - {T_A}_1 + {k_1}( {{L_1} - {L_2}})}))}}
{{( {{N_A}{{_1}_1} - {N_A}{{_2}_1}})( {{N_A}{{_5}_2} - {N_A}{{_6}_2}}) - ( {{N_A}{{_1}_2} - {N_A}{{_2}_2}})( {{N_A}{{_5}_1} - {N_A}{{_6}_1}})}}\\
{E_{A4}} = {T_A}_4 + \frac{{N_A}{{_4}_2}( {( {{N_A}{{_1}_1} - {N_A}{{_2}_1}})( {{T_A}_6 - {T_A}_5 + {k_5}( {{L_5} - {L_6}})}))}}
{{( {{N_A}{{_1}_1} - {N_A}{{_2}_1}})( {{N_A}{{_5}_2} - {N_A}{{_6}_2}}) - ( {{N_A}{{_1}_2} - {N_A}{{_2}_2}})( {{N_A}{{_5}_1} - {N_A}{{_6}_1}})}}\\
{\quad\quad\;\;}- \frac{{N_A}{{_4}_2}({( {{N_A}{{_5}_1} - {N_A}{{_6}_1}})( {{T_A}_2 - {T_A}_1 + {k_1}( {{L_1} - {L_2}})}))}}
{{( {{N_A}{{_1}_1} - {N_A}{{_2}_1}})( {{N_A}{{_5}_2} - {N_A}{{_6}_2}}) - ( {{N_A}{{_1}_2} - {N_A}{{_2}_2}})( {{N_A}{{_5}_1} - {N_A}{{_6}_1}})}}\\
{\quad\quad\;\;}- \frac{{N_A}{{_4}_1}( {( {{N_A}{{_1}_2} - {N_A}{{_2}_2}})( {{T_A}_6 - {T_A}_5 + {k_5}( {{L_5} - {L_6}})}))}}
{{( {{N_A}{{_1}_1} - {N_A}{{_2}_1}})( {{N_A}{{_5}_2} - {N_A}{{_6}_2}}) - ( {{N_A}{{_1}_2} - {N_A}{{_2}_2}})( {{N_A}{{_5}_1} - {N_A}{{_6}_1}})}}\\
{\quad\quad\;\;}+ \frac{{N_A}{{_4}_1}({( {{N_A}{{_5}_2} - {N_A}{{_6}_2}})( {{T_A}_2 - {T_A}_1 + {k_1}( {{L_1} - {L_2}})}))}}
{{( {{N_A}{{_1}_1} - {N_A}{{_2}_1}})( {{N_A}{{_5}_2} - {N_A}{{_6}_2}}) - ( {{N_A}{{_1}_2} - {N_A}{{_2}_2}})( {{N_A}{{_5}_1} - {N_A}{{_6}_1}})}}\\
{E_{A5}} = {T_A}_5 + \frac{{N_A}{{_5}_2}( {( {{N_A}{{_1}_1} - {N_A}{{_2}_1}})( {{T_A}_6 - {T_A}_5 + {k_5}( {{L_5} - {L_6}})}))}}
{{( {{N_A}{{_1}_1} - {N_A}{{_2}_1}})( {{N_A}{{_5}_2} - {N_A}{{_6}_2}}) - ( {{N_A}{{_1}_2} - {N_A}{{_2}_2}})( {{N_A}{{_5}_1} - {N_A}{{_6}_1}})}}\\
{\quad\quad\;\;}- \frac{{N_A}{{_5}_2}({( {{N_A}{{_5}_1} - {N_A}{{_6}_1}})( {{T_A}_2 - {T_A}_1 + {k_1}( {{L_1} - {L_2}})}))}}
{{( {{N_A}{{_1}_1} - {N_A}{{_2}_1}})( {{N_A}{{_5}_2} - {N_A}{{_6}_2}}) - ( {{N_A}{{_1}_2} - {N_A}{{_2}_2}})( {{N_A}{{_5}_1} - {N_A}{{_6}_1}})}}\\
{\quad\quad\;\;}- \frac{{N_A}{{_5}_1}( {( {{N_A}{{_1}_2} - {N_A}{{_2}_2}})( {{T_A}_6 - {T_A}_5 + {k_5}( {{L_5} - {L_6}})}))}}
{{( {{N_A}{{_1}_1} - {N_A}{{_2}_1}})( {{N_A}{{_5}_2} - {N_A}{{_6}_2}}) - ( {{N_A}{{_1}_2} - {N_A}{{_2}_2}})( {{N_A}{{_5}_1} - {N_A}{{_6}_1}})}}\\
{\quad\quad\;\;}+ \frac{{N_A}{{_5}_1}({( {{N_A}{{_5}_2} - {N_A}{{_6}_2}})( {{T_A}_2 - {T_A}_1 + {k_1}( {{L_1} - {L_2}})}))}}
{{( {{N_A}{{_1}_1} - {N_A}{{_2}_1}})( {{N_A}{{_5}_2} - {N_A}{{_6}_2}}) - ( {{N_A}{{_1}_2} - {N_A}{{_2}_2}})( {{N_A}{{_5}_1} - {N_A}{{_6}_1}})}}\\
{E_{A6}} = {T_A}_6 + \frac{{N_A}{{_6}_2}( {( {{N_A}{{_1}_1} - {N_A}{{_2}_1}})( {{T_A}_6 - {T_A}_5 + {k_5}( {{L_5} - {L_6}})}))}}
{{( {{N_A}{{_1}_1} - {N_A}{{_2}_1}})( {{N_A}{{_5}_2} - {N_A}{{_6}_2}}) - ( {{N_A}{{_1}_2} - {N_A}{{_2}_2}})( {{N_A}{{_5}_1} - {N_A}{{_6}_1}})}}\\
{\quad\quad\;\;}- \frac{{N_A}{{_6}_2}({( {{N_A}{{_5}_1} - {N_A}{{_6}_1}})( {{T_A}_2 - {T_A}_1 + {k_1}( {{L_1} - {L_2}})}))}}
{{( {{N_A}{{_1}_1} - {N_A}{{_2}_1}})( {{N_A}{{_5}_2} - {N_A}{{_6}_2}}) - ( {{N_A}{{_1}_2} - {N_A}{{_2}_2}})( {{N_A}{{_5}_1} - {N_A}{{_6}_1}})}}\\
{\quad\quad\;\;}- \frac{{N_A}{{_6}_1}( {( {{N_A}{{_1}_2} - {N_A}{{_2}_2}})( {{T_A}_6 - {T_A}_5 + {k_5}( {{L_5} - {L_6}})}))}}
{{( {{N_A}{{_1}_1} - {N_A}{{_2}_1}})( {{N_A}{{_5}_2} - {N_A}{{_6}_2}}) - ( {{N_A}{{_1}_2} - {N_A}{{_2}_2}})( {{N_A}{{_5}_1} - {N_A}{{_6}_1}})}}\\
{\quad\quad\;\;}+ \frac{{N_A}{{_6}_1}({( {{N_A}{{_5}_2} - {N_A}{{_6}_2}})( {{T_A}_2 - {T_A}_1 + {k_1}( {{L_1} - {L_2}})}))}}
{{( {{N_A}{{_1}_1} - {N_A}{{_2}_1}})( {{N_A}{{_5}_2} - {N_A}{{_6}_2}}) - ( {{N_A}{{_1}_2} - {N_A}{{_2}_2}})( {{N_A}{{_5}_1} - {N_A}{{_6}_1}})}}
\end{array}\]

\bibliographystyle{asmems4}

\bibliography{J5_Bibliography}



\end{document}